\PassOptionsToPackage{capitalize}{cleveref}
\documentclass[10pt,twocolumn,letterpaper]{article}

\usepackage[pagenumbers]{wacv}

\definecolor{wacvblue}{rgb}{0.21,0.49,0.74}
\usepackage{xr-hyper}
\usepackage[pagebackref,breaklinks,colorlinks,allcolors=wacvblue]{hyperref}
\usepackage{cleveref}
\usepackage[table]{xcolor}

\definecolor{pearFour}{HTML}{588F27}
\definecolor{pearFith}{HTML}{ECF0F1}
\definecolor{pearDark}{HTML}{2980B9}
\definecolor{pearDarker}{HTML}{1D2DEC}
\definecolor{pearLight}{HTML}{F7E0D5}

\definecolor{eva01purple}{RGB}{168,119,200}
\definecolor{eva02red}{RGB}{236,35,35}

\usepackage{url}
\usepackage{nicefrac}
\usepackage{lipsum}
\usepackage{fancyhdr}
\usepackage{appendix}
\usepackage{afterpage}
\usepackage{comment}

\usepackage{graphicx}
\usepackage{caption}
\usepackage{wrapfig}
\usepackage{float}
\usepackage{soul}

\usepackage{booktabs}
\usepackage{multirow}
\usepackage{threeparttable}
\usepackage{arydshln}

\usepackage{amsmath}
\usepackage{amssymb}
\usepackage{amsfonts}
\usepackage{bm}
\usepackage{dsfont}

\newcommand{\cmark}{\ding{51}}%
\newcommand{\xmark}{\ding{55}}%

\usepackage{algorithm}
\usepackage{algorithmic}
\usepackage{listings}
\usepackage{verbatim}
\usepackage{comment}

\usepackage{etoolbox}
\makeatletter
\AfterEndEnvironment{algorithm}{\let\@algcomment\relax}
\AtEndEnvironment{algorithm}{\kern2pt\hrule\relax\vskip3pt\@algcomment}
\let\@algcomment\relax
\newcommand\algcomment[1]{\def\@algcomment{\footnotesize#1}}
\renewcommand\fs@ruled{\def\@fs@cfont{\bfseries}\let\@fs@capt\floatc@ruled
  \def\@fs@pre{\hrule height.8pt depth0pt \kern2pt}%
  \def\@fs@post{}%
  \def\@fs@mid{\kern2pt\hrule\kern2pt}%
  \let\@fs@iftopcapt\iftrue}
\makeatother

\usepackage{pifont}
\usepackage{textcomp}
\usepackage{relsize}
\usepackage{xspace}

\usepackage{cancel}

\usepackage{tikz}
\usepackage{placeins}
\usetikzlibrary{decorations.pathreplacing, positioning}

\newcommand{\bx}{\mathbf{x}}
\newcommand{\by}{\mathbf{y}}
\newcommand{\bz}{\mathbf{z}}
\newcommand{\bmu}{\boldsymbol{\mu}}
\newcommand{\bsig}{\boldsymbol{\sigma}}
\newcommand{\beps}{\boldsymbol{\varepsilon}}

\newcommand{\sx}{\mathbf{s}_{\mathrm{x}}}
\newcommand{\sy}{\mathbf{s}_{\mathrm{y}}}
\newcommand{\hsy}{\mathbf{\hat{s}}_{\mathrm{y}}}

\newcommand{\prior}{p_\theta(\bz \mid \sx)}
\newcommand{\post}{q_\psi(\bz \mid \sx,\, \sy)}

\newcommand{\Lpred}{\mathcal{L}_{\mathrm{pred}}}
\newcommand{\Lreg}{\mathcal{L}_{\mathrm{reg}}}
\newcommand{\Ltotal}{\mathcal{L}}
\newcommand{\KL}{D_{\mathrm{KL}}}

\usepackage{amsthm}
\usepackage{tcolorbox}
\tcbuselibrary{theorems,skins,breakable}

\newtcbtheorem
  [
  crefname={detail}{detail}]
  {detail}
  {Experiment Details}
  {%
    fontupper=\small,
    colback=orange!5,
    colframe=orange!35!black,
    fonttitle=\bfseries,
    boxsep=1pt,
    left=1.5mm,
    right=1.5mm,
    top=2mm,
    bottom=1mm,
  }
  {detail}
\newtcbtheorem
  [
  crefname={def.}{def.}]
  {definition}
  {Definition}
  {%
    fontupper=\small,
    colback=green!5,
    colframe=green!35!black,
    fonttitle=\bfseries,
    boxsep=1pt,
    left=1.5mm,
    right=1.5mm,
    top=2mm,
    bottom=1mm,
  }
  {def}
\newtcbtheorem
  [
  crefname={thm.}{thms.}]
  {theorem}
  {Theorem}
  {%
    fontupper=\small,
    colback=red!5,
    colframe=red!35!black,
    fonttitle=\bfseries,
    boxsep=1pt,
    left=1.5mm,
    right=1.5mm,
    top=2mm,
    bottom=1mm,
  }
  {theorem}
\newtcbtheorem
  [
  crefname={lemma.}{lemmas.}]
  {lemma}
  {Lemma}
  {%
    fontupper=\small,
    colback=blue!3,
    colframe=blue!35!black,
    fonttitle=\bfseries,
    boxsep=1pt,
    left=1.5mm,
    right=1.5mm,
    top=2mm,
    bottom=1mm,
  }
  {lemma}
\newtcbtheorem
  [
  crefname={prop.}{props.}]
  {proposition}
  {Proposition}
  {%
    breakable,
    enhanced,
    fontupper=\small,
    colback=red!5,
    colframe=red!35!black,
    fonttitle=\bfseries,
    boxsep=1pt,
    left=1.5mm,
    right=1.5mm,
    top=2mm,
    bottom=1mm,
  }
  {proposition}
\newtcbtheorem
  [
  crefname={cor.}{corrs.}]
  {corollary}
  {Corollary}
  {%
    breakable,
    enhanced,
    fontupper=\small,
    colback=red!5,
    colframe=red!35!black,
    fonttitle=\bfseries,
    boxsep=1pt,
    left=1.5mm,
    right=1.5mm,
    top=2mm,
    bottom=1mm,
  }
  {corollary}

\newcommand{\model}{S-JEPA\xspace}

\crefname{lemma}{Lemma}{Lemmas}
\Crefname{lemma}{Lemma}{Lemmas}

\def\wacvPaperID{****}
\def\confName{WACV}
\def\confYear{2027}

\title{
A Stochastic Latent Variable Generalization of \\Joint-Embedding Predictive Architectures
}

\author{
Nikos Giakoumoglou$^{1}$\thanks{Corresponding author: \texttt{nikos@imperial.ac.uk}}
\qquad
Paschalis Giakoumoglou$^{2}$
\qquad
Tania Stathaki$^{1}$\\[0.5em]
$^{1}$Imperial College London \qquad
$^{2}$Information Technologies Institute, CERTH\vspace{.5em}\\
{Code: \url{https://github.com/giakoumoglou/sjepa}}\\
}

\begin{document}
\maketitle

\begin{abstract}
Joint-Embedding Predictive Architectures (JEPAs) predict the latent embedding of a target signal from a context signal, with an auxiliary variable $\bz$ conditioning the predictor. Despite its prominent role in the original JEPA blueprint, the content of $\bz$ has been carried by a fixed deterministic token in every prior instantiation, collapsing the genuinely multi-modal distribution over plausible completions to a point estimate. We present \model, which realises $\bz$ as a proper stochastic latent variable for masked image prediction: a learned posterior infers $\bz$ from context and target during training, while a learned context-conditioned prior approximates the distribution of plausible completions at inference, with no decoders and no reconstruction. In its deterministic limit the objective reduces to the I-JEPA loss in functional form, generalising that objective rather than strictly containing the I-JEPA model. Our central finding is that \emph{stochasticity adds capabilities, error-correlated predictive uncertainty and diverse completions, at no cost to representation quality}: under matched architectures and schedules \model improves over the deterministic I-JEPA baseline and exceeds LeJEPA at matched architecture, while additionally producing spatially structured uncertainty maps and semantically diverse completions, neither of which the deterministic predictor produces.
\end{abstract}

\section{Introduction}
\label{sec:introduction}

Self-supervised representation learning has emerged as a promising paradigm for advancing machine intelligence without explicit supervision~\citep{radford2018gpt,chen2020simclr}, producing representations that rival or exceed fully-supervised training on many downstream tasks~\citep{giakoumoglou2024review,chen2020simclr,grill2020byol,he2019moco,caron2021dino,he2021mae,giakoumoglou2026vitamins}. \citet{lecun2022path} introduced \emph{Joint-Embedding Predictive Architectures} (JEPAs) as a step toward a ``\emph{world model}''. Unlike generative models~\citep{van2016pixel,parmar2018imagetransformer,he2021mae,bao2021beit} that reconstruct raw inputs, JEPAs predict in \emph{latent} space: given a context signal, the predictor recovers the \emph{representation} of a target signal, not raw pixels. \citet{assran2023ijepa} instantiated this for vision, showing that predicting masked patch representations from a context block yields semantic embeddings without hand-crafted view augmentations, outperforming reconstruction-based baselines~\citep{he2021mae,bao2021beit,chen2022cae} on linear probing; the framework has since been extended to video~\citep{bardes2024vjepa,assran2025vjepa2,murlabadia2026vjepa2_1}, language~\citep{huang2026llmjepa}, and vision-language grounding~\citep{chen2026vljepa}, and endowed with theoretically grounded collapse-prevention objectives~\citep{balestriero2025lejepa,kuang2026lpjepa}.

\begin{figure}[!t]
    \centering
    \includegraphics[width=0.9\linewidth]{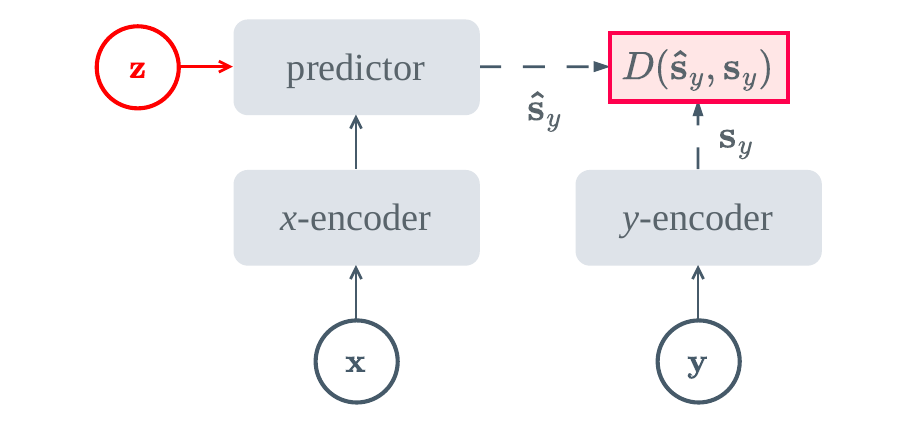}
    \caption{{\bf JEPAs} are trained to predict the representation of an input $\by$ from the representation of another input $\bx$. The additional variable $\bz$ provides the predictor with information about the transformation that computes $\by$ from $\bx$.}
    \label{fig:jepa}
\end{figure}

Concretely, a JEPA comprises a context $x$-encoder $f_x$ and a target $y$-encoder $f_y$, producing $\sx = f_x(\bx)$ and $\sy = f_y(\by)$, plus a predictor $g$ that estimates $\hsy = g(\sx, \bz)$ (\Cref{fig:jepa}). Central to LeCun's blueprint~\citep{lecun2022path,dawid2023introduction} is the \emph{auxiliary variable} $\bz$, carrying the residual information the context lacks, \eg scene-level factors, global style, or the semantic content of the masked region. Yet $\bz$ is deterministic in every instantiation: a shared learnable mask token with positional embedding in I-JEPA~\citep{assran2023ijepa}, and its spatiotemporal analogue in V-JEPA~\citep{bardes2024vjepa}. Stochasticity has entered only the \emph{positional} component, by sampling the mask-token position~\citep{bar2024stochastic} or conditioning the encoders on window positions~\citep{littwin2024enhancingjepas}; neither models a distribution over what the masked region contains, the role $\bz$ carries in the blueprint.

This gap matters because masked prediction is intrinsically ambiguous. Given ``\emph{pavement}'', a ``\emph{parked car}'', and ``\emph{open sky}'', the hidden region could equally plausibly contain a ``\emph{fire hydrant}'', a ``\emph{pedestrian}'', or a ``\emph{traffic signal}''. Even with perfect context representations the distribution over semantically valid completions is genuinely multi-modal~\citep{lecun2022path}, and it remains so once the location of the target is known. A predictor that collapses this distribution to a point estimate discards the space of plausible completions and offers no signal indicating where the context under-determines the target.

We therefore realise $\bz$ as a proper stochastic latent variable over the content of the masked region, fulfilling the intent of LeCun's original formulation~\citep{lecun2022path,dawid2023introduction}, replacing the deterministic predictor with one built on a Conditional Variational Autoencoder (CVAE)~\citep{sohn2015cvae,kingma2013vae} operating \emph{entirely in representation space}. A learned posterior $\post$ infers $\bz$ from context and target during training, while a learned prior $\prior$ anticipates plausible completions from context alone and is used at inference. Training maximises a variational lower bound (ELBO) on the log-likelihood of $\sy$ given $\sx$, decomposing into a prediction term and a Kullback-Leibler (KL) divergence regulariser (\Cref{def:sjepa}), with the reparameterisation trick~\citep{kingma2013vae} enabling end-to-end gradient flow through the inference network. Collapsing $\bz$ to a deterministic code with no access to the target reduces the prediction term to the JEPA loss of \citet{assran2023ijepa} in functional form, while the KL regulariser is precisely what prevents the target-conditioned posterior from short-cutting prediction; our formulation is thus a probabilistic generalisation of the I-JEPA objective rather than a strict superset of the I-JEPA model.

Our \textbf{contributions} are as follows.
\textbf{(i)} We realise the auxiliary latent variable $\bz$ in JEPA as a stochastic variable over the \emph{content} of the masked region for spatial masked image prediction, using a learned context-conditioned prior and no decoders or reconstruction, directly addressing the intent of LeCun's original JEPA blueprint~\citep{lecun2022path}; prior stochastic treatments of $\bz$ act on its positional component~\citep{bar2024stochastic,littwin2024enhancingjepas}, and concurrent variational JEPA work targets sequential control, tabular data, or symmetric view modelling (\Cref{sec:related}), settings distinct from ours.
\textbf{(ii)} We derive a principled CVAE-style ELBO for JEPA-style masked prediction, with a learned conditional prior and posterior over $\bz$, operating entirely in latent space with no observation-space reconstruction.
\textbf{(iii)} We show that the proposed objective recovers the original JEPA loss in functional form as a deterministic special case, providing a clean interpolation between point-estimate and distributional prediction.
\textbf{(iv)} We demonstrate that \emph{stochasticity adds capabilities, error-correlated predictive uncertainty (\Cref{fig:uncertainty}) and diverse completions (\Cref{fig:rcdm}), at no cost to representation quality}: \model improves over I-JEPA~\citep{assran2023ijepa} under matched architectures and schedules and exceeds LeJEPA~\citep{balestriero2025lejepa} at matched architecture on a longer schedule, while additionally producing an error-correlated uncertainty signal and semantically diverse completions that the deterministic predictor does not produce.
\section{Related Work}
\label{sec:related}

\subsection{Joint-Embedding Predictive Architectures}
\label{subsec:related-jepa}

JEPAs~\citep{lecun2022path,dawid2023introduction} predict missing information from context in \emph{latent} rather than pixel space. I-JEPA~\citep{assran2023ijepa} is the canonical image instantiation: a context encoder processes visible patches and a narrow predictor, conditioned on positional mask tokens, predicts the representations of target blocks computed by an EMA target encoder. The framework has been extended to video~\citep{bardes2024vjepa,assran2025vjepa2,murlabadia2026vjepa2_1}, joint motion and content~\citep{bardes2023mcjepa}, language~\citep{huang2026llmjepa}, vision-language grounding~\citep{chen2026vljepa}, and heterogeneous tabular data~\citep{thimonier2024tjepa}, and connected to contrastive learning~\citep{mo2024cjepa}. In all of these, the auxiliary variable $\bz$ is treated as a fixed deterministic signal. Two works make part of it stochastic, both acting on its positional component: \citet{bar2024stochastic} draw the mask-token positional embedding from a Gaussian with a learned covariance shared across tokens and images, and \citet{littwin2024enhancingjepas} condition each encoder on the positions of the opposing window. Neither introduces a posterior over the target, a context-conditioned prior, or a distribution over the content of the masked region. Since the two mechanisms act on different arguments of $g(\sx, \bz)$ they are complementary to ours rather than competing; we compare in Appendix~\ref{sec:app-discussion}.

\subsection{Collapse Prevention and Distributional Regularisation}
\label{subsec:related-collapse}

Standard JEPA implementations rely on EMA updates~\citep{grill2020byol} of the target encoder~\citep{assran2023ijepa,bardes2024vjepa}, which prevent trivial solutions but offer no explicit control over the distributional properties of embeddings~\citep{giakoumoglou2026three}. VICReg~\citep{bardes2021vicreg} penalises feature-wise variance collapse and cross-dimension covariance, and \citet{drozdov2024video} apply variance-covariance penalties within JEPA; \citet{balestriero2025lejepa} prove that an isotropic Gaussian is the minimax-optimal embedding distribution for downstream probing and enforce it via SIGReg, eliminating EMA and all training heuristics, and \citet{kuang2026lpjepa} extend this with a sliced two-sample loss for explicit sparsity control. In \model, target-information leakage through the predictor is controlled by the KL term of the CVAE-style ELBO, which acts as an information bottleneck on $\bz$ and prevents it from becoming a degenerate free variable. This addresses a distinct failure mode acting on a distinct object: the KL bounds the information $\bz$ carries about $\sy$ (Lemma~\ref{lemma:ib}) but places no constraint on the encoder outputs, so the EMA target encoder is retained for the context-target pair, consistent with I-JEPA~\citep{assran2023ijepa}, and we do not claim it is made redundant.

\subsection{Probabilistic Prediction in Latent Space}
\label{subsec:related-probabilistic}

The VAE~\citep{kingma2013vae} and CVAE~\citep{sohn2015cvae} provide the ELBO and reparameterisation machinery we build on, while world-model research has emphasised stochastic latent dynamics for planning~\citep{ha2018worldmodels}. The prior-posterior asymmetry we adopt has a precedent there: \citet{hafner2019learning} pair a learned transition prior with an observation-conditioned posterior precisely to represent multiple futures, but close the loop through a pixel decoder (Appendix~\ref{subsec:disc-rssm}); distributional embeddings in retrieval~\citep{oh2019modeling,chun2021probabilistic} attach uncertainty to an observed input rather than a withheld region (Appendix~\ref{subsec:disc-probemb}). Two concurrent works introduce probabilistic objectives into JEPAs but differ substantially from ours: \citet{huang2026vjepa} target sequential prediction over future latent states with a fixed Gaussian prior, and \citet{gogl2026varjepa} formulate JEPA as a full generative model with observation-space decoders and reconstruction terms. \model differs on three axes: \textbf{(i)} it addresses spatial masked image prediction, \textbf{(ii)} it employs a context-conditioned learned prior $\prior$, and \textbf{(iii)} it contains \emph{no decoders or reconstruction objective}, the CVAE operating purely in representation space and preserving the non-generative character of JEPA~\citep{lecun2022path,assran2023ijepa}. A detailed analytical comparison is given in Appendix~\ref{sec:app-discussion}.
\section{Methodology}
\label{sec:method}

We first review the JEPA framework and its deterministic predictor (\Cref{subsec:preliminaries}), then derive the \model objective by introducing $\bz$ as a stochastic latent variable via a CVAE operating in representation space (\Cref{subsec:formulation}).

\subsection{Preliminaries: Joint-Embedding Predictive Architectures}
\label{subsec:preliminaries}

In JEPAs~\citep{lecun2022path}, a system learns to predict the \emph{representation} of one view of the data from the representation of another, rather than reconstructing the raw input signal (\Cref{fig:jepa}). A JEPA consists of a context $x$-encoder $f_x$, a target $y$-encoder $f_y$, and a predictor $g$. Given a context observation $\bx$ and a target observation $\by$, the model computes $\sx = f_x(\bx)$, $\sy = f_y(\by)$, and $\hsy = g(\sx, \bz)$, where $\bz$ is an auxiliary variable that provides the predictor with additional information to facilitate prediction. The training objective minimises the distance between the predicted and actual target representations in embedding space~\citep{assran2023ijepa}.

\begin{definition}{I-JEPA}{jepa}
\begin{equation}
    \mathcal{L}_{\mathrm{JEPA}} = \mathbb{E}\!\left[ \| g(\sx, \bz) - \mathrm{sg}(f_y(\by)) \|_2^2 \right],
    \label{eq:jepa-loss}
    \tag{I-JEPA}
\end{equation}
where $\mathrm{sg}(\cdot)$ denotes the stop-gradient operation, which blocks gradients from flowing into the target encoder.
\end{definition}

\noindent 
It is the deterministic treatment of the content of $\bz$ in \Cref{def:jepa}, described in \Cref{sec:introduction}, that we now relax.

\subsection{Stochastic Prediction in Representation Space}
\label{subsec:formulation}

\begin{figure}[!t]
    \centering
    \includegraphics[width=0.9\linewidth]{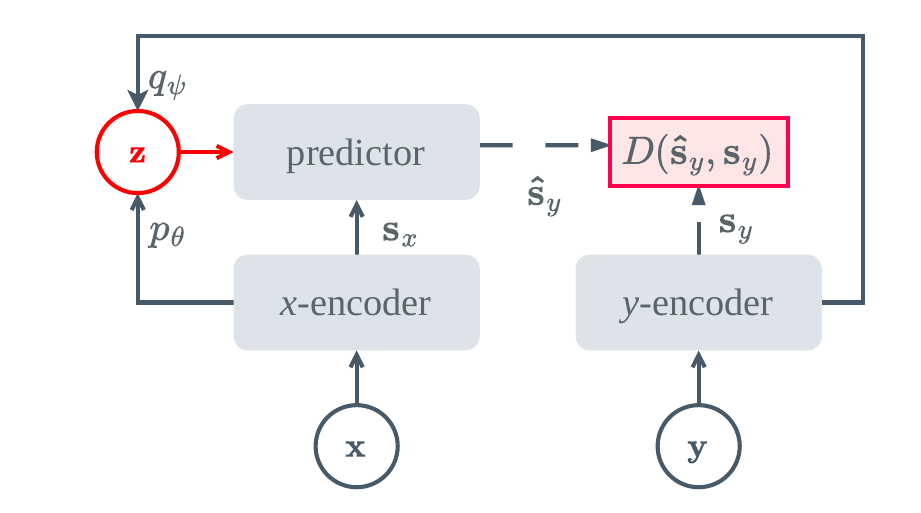}
    \caption{{\bf \model overview.} The posterior network $\post$ infers $\bz$ from both context and target representations; the prior network $\prior$ infers $\bz$ from context alone. The sampled $\bz$ is injected into the predictor $g_\phi$ to produce $\hsy$. No arrows return to observation space: \model is non-generative by design.}
    \label{fig:sjepa}
\end{figure}

We began by asking a simple structural question: \emph{if the predictive embedding steps in JEPA were interpreted as the variational posteriors of a coupled VAE~\citep{kingma2013vae,sohn2015cvae}, what probabilistic model would give rise to those posteriors}? This viewpoint naturally leads to a novel reinterpretation of JEPA within a probabilistic latent-variable framework. Concretely, we replace the deterministic $\bz$ with a stochastic latent variable and introduce a learned prior and posterior over it, both operating purely in representation space, preserving the non-generative character of JEPA throughout. We refer to the resulting model as \model (\textbf{\underline{S}}tochastic \textbf{\underline{J}}oint-\textbf{\underline{E}}mbedding \textbf{\underline{P}}redictive \textbf{\underline{A}}rchitecture). The overall structure is illustrated in \Cref{fig:sjepa}.

\paragraph{\bf Objective and factorisation.} Our goal is to model the conditional distribution $p(\sy \mid \sx)$ in representation space, \ie, the distribution over target patch representations given the context representation of the same image. Even with full context, the masked regions retain genuine uncertainty due to occlusion, texture variability, and semantic ambiguity~\citep{lecun2022path}. We introduce $\bz \in \mathbb{R}^{d_z}$ as a per-image latent variable that accounts for this residual uncertainty, giving the joint:
\begin{equation}
    p_{\phi,\theta}(\sy, \bz \mid \sx) = p_\phi(\sy \mid \sx, \bz)\, \prior,
    \label{eq:joint}
\end{equation}
where $\prior$ is a learned context-conditioned prior parameterised by $\theta$, and $p_\phi(\sy \mid \sx, \bz)$ is the likelihood defined through the predictor $g_\phi$. The marginal likelihood of the target representation given context is:
\begin{equation}
    p(\sy \mid \sx) = \int p_\phi(\sy \mid \sx, \bz)\, \prior\, d\bz,
    \label{eq:marginal}
\end{equation}
which is intractable for a learned neural predictor $g_\phi$. We instead maximise a variational lower bound.

\paragraph{\bf Context-conditioned prior.} The prior $\prior$ conditions on the context representation $\sx$ alone, and must learn to anticipate the distribution of plausible target completions from visible patches without access to the target. We parameterise it as a diagonal Gaussian:
\begin{equation}
    \prior = \mathcal{N}\!\left(\bmu_\theta(\sx),\, \mathrm{diag}(\bsig_\theta^2(\sx))\right),
    \label{eq:prior}
\end{equation}
where $\bmu_\theta$ and $\bsig_\theta$ are learned functions of $\sx$. At inference, $\bz$ is sampled from this prior since the target representation is unavailable.

\paragraph{\bf Approximate posterior.} During training, we introduce an approximate posterior $\post$ that additionally conditions on the target representation $\sy$, allowing it to infer the latent code that best explains the observed target given the context:
\begin{equation}
    \post = \mathcal{N}\!\left(\bmu_\psi(\sx, \sy),\, \mathrm{diag}(\bsig_\psi^2(\sx, \sy))\right).
    \label{eq:posterior}
\end{equation}
At inference the posterior is unavailable and $\bz$ is drawn from the prior $\prior$ instead. The prior must therefore learn to match the aggregate distribution of posteriors across the training data, driving it to internalise the statistics of plausible completions conditioned on context alone (see Appendix~\ref{subsec:remark-prior}).

\paragraph{\bf Evidence lower bound.} Introducing the approximate posterior and applying Jensen's inequality to the concave $\log$:
\begin{align}
    &\log p(\sy \mid \sx) \notag \\
    &= \log \int p_\phi(\sy \mid \sx, \bz)\, \prior\, d\bz \notag \\
    &= \log \int \post \cdot \frac{p_\phi(\sy \mid \sx, \bz)\, \prior}{\post}\, d\bz \notag \\
    &= \log\, \mathbb{E}_{q_\psi}\!\left[\frac{p_\phi(\sy \mid \sx, \bz)\, \prior}{\post}\right] \notag \\
    &\geq \mathbb{E}_{q_\psi}\!\left[\log p_\phi(\sy \mid \sx, \bz)\right] - \KL\!\left(q_\psi \;\|\; p_\theta\right).
    \label{eq:elbo}
\end{align}
This is the CVAE ELBO~\citep{sohn2015cvae} in representation space (Prop.~\ref{proposition:elbo}, Appendix~\ref{subsec:proof-elbo}). The first term drives accurate prediction of $\sy$ given $\sx$ and $\bz$; the second term regularises $\bz$ toward the context-conditioned prior, preventing the posterior from encoding information inaccessible at inference. The bound is tight if and only if $\post$ equals the true posterior $p(\bz \mid \sx, \sy)$ (see Appendix~\ref{subsec:remark-gap}).

\paragraph{\bf Gaussian likelihood and MSE reduction.} We assume a fixed-variance isotropic Gaussian likelihood:
\begin{equation}
    p_\phi(\sy \mid \sx, \bz) = \mathcal{N}\!\left(\sy;\, g_\phi(\sx, \bz),\, \sigma^2 \mathbf{I}\right),
    \label{eq:gaussian-likelihood}
\end{equation}
where $\sigma^2 > 0$ is a fixed observation-noise parameter. Expanding the log-density and taking the expectation under $q_\psi$ yields (Prop.~\ref{proposition:gaussian}, Appendix~\ref{subsec:proof-gaussian}):
\begin{equation}
    -\mathbb{E}_{q_\psi}\!\left[\log p_\phi(\sy \mid \sx, \bz)\right] = \frac{1}{2\sigma^2}\,\mathbb{E}_{q_\psi}\!\left[\|\sy - g_\phi(\sx, \bz)\|^2\right] + C,
    \label{eq:mse_reduction}
\end{equation}
where $C = \frac{d_y}{2}\log(2\pi\sigma^2)$ is a constant independent of $\phi$, $\psi$, $\theta$ (here $d_y$ denotes the dimensionality of $\sy$). Minimising the negative ELBO with respect to $(\phi, \psi, \theta)$ is therefore equivalent to minimising the scaled MSE plus the KL term. The KL divergence between the two diagonal Gaussians is computed in closed form (Prop.~\ref{proposition:kl}, Appendix~\ref{subsec:proof-kl}):
\begin{align}
    &\KL\!\left(q_\psi \;\|\; p_\theta\right) = \notag \\
    &\quad \frac{1}{2} \sum_{k=1}^{d_z} \left( \log \frac{\sigma_{\theta,k}^2}{\sigma_{\psi,k}^2} + \frac{\sigma_{\psi,k}^2 + (\mu_{\psi,k} - \mu_{\theta,k})^2}{\sigma_{\theta,k}^2} - 1 \right).
    \label{eq:kl}
\end{align}
Sampling is performed via the reparameterisation trick~\citep{kingma2013vae}, $\bz = \bmu_\psi + \bsig_\psi \odot \beps$ with $\beps \sim \mathcal{N}(\mathbf{0}, \mathbf{I})$, which allows gradient flow through the stochastic node. Absorbing the constant $1/(2\sigma^2)$ into a single hyperparameter $\beta > 0$ that controls the relative strength of the KL regulariser (Appendix~\ref{subsec:remark-beta}), we obtain the final \model training objective:

\begin{definition}{Stochastic JEPA}{sjepa}
\begin{align}
    \Ltotal &= \underbrace{\mathbb{E}_{q_\psi}\!\left[\| \mathrm{sg}(\sy) - g_\phi(\sx, \bz) \|^2\right]}_{\Lpred} \notag \\
    &\quad + \underbrace{\beta \cdot \KL\!\left(\post \;\|\; \prior\right)}_{\Lreg},
    \label{eq:total-loss}
    \tag{\model}
\end{align}
where $\mathrm{sg}(\cdot)$ is the stop-gradient of \Cref{def:jepa}, retained on the EMA target encoder, and $\beta > 0$ controls the information bottleneck on $\bz$.\footnote{The posterior $\post$ conditions on $\mathrm{sg}(\sy)$, so gradients reach $\psi$ but not $f_y$.} Setting $\beta = 2\sigma^2$ recovers the exact (scaled) negative ELBO; treating $\beta$ as a free hyperparameter yields a $\beta$-VAE-style objective~\citep{higgins2017betavae}.
\end{definition}

\noindent When $\bz$ is collapsed to a deterministic code that does not access the target, $\Ltotal$ reduces to the I-JEPA objective of \Cref{def:jepa} in functional form, recovering the deterministic baseline as a special case of \model (Thm.~\ref{theorem:ijepa}, Appendix~\ref{subsec:proof-ijepa}); the I-JEPA mask token is target-independent whereas $\bmu_\psi$ is not, so the recovery is of the loss form rather than the specific parameterisation.
\section{Implementation Details}
\label{sec:implementation}

We use a vision transformer~\citep{dosovitskiy2020vit} with patch size $16{\times}16$ at resolution $224{\times}224$ in two configurations: ViT-B/16 and ViT-L/16. The target encoder is an EMA copy of the context encoder with momentum linearly scheduled from 0.996 to 1.0. The predictor is a standard Transformer of depth 6 for ViT-B and 12 for ViT-L, embedding dimension 384, 12 heads, and MLP ratio 4; context tokens are projected into the predictor dimension, $\bz$ is prepended as a prefix token, and learnable mask tokens are appended for each target position before a final projection back to the encoder dimension. Both the prior $\prior$ and posterior $\post$ are 2-layer GELU MLPs mapping to a diagonal Gaussian over $\mathbb{R}^{128}$. We use the \texttt{multi-block} masking strategy of~\citet{assran2023ijepa} for \model: a single context block with scale $(0.85,1.0)$ and unit aspect ratio, and $4$ target blocks with scale $(0.15,0.20)$ and aspect ratio $(0.75,1.5)$, with any overlap between the context and target blocks removed; we adopt \texttt{multi-block} to match I-JEPA~\citep{assran2023ijepa}, holding the pretext task fixed across the two predictors. We train with AdamW~\citep{loshchilov2017adamw}, batch size 2048 for ViT-B/16 and 1024 for ViT-L/16, on 4 A100 GPUs with \texttt{bfloat16}, warming the learning rate from $2{\times}10^{-4}$ to $10^{-3}$ over 40 epochs then cosine-decaying to $10^{-6}$, with weight decay cosine-scheduled from 0.04 to 0.4, for 600 epochs on ImageNet-1K~\citep{deng2009imagenet}. Pretraining and evaluation details are in Appendix~\ref{sec:app-implementation}; all ablations are in Appendix~\ref{sec:app-ablation}.

\section{Experiments}
\label{sec:experiments}

We evaluate \model on image classification (\Cref{subsec:classification}), low-level vision (\Cref{subsec:lowlevel}), qualitative analysis of encoder and predictor behaviour (\Cref{subsec:encoder-viz,subsec:predictor-viz}), and the computational cost of stochasticity (\Cref{subsec:cost}). Main-paper benchmark results use a single run; appendix ablations average three pretraining seeds.

\subsection{Image Classification}
\label{subsec:classification}

\begin{table}[!t]
\centering
\caption{{\bf ImageNet linear evaluation.} Top-1 linear probing accuracy on frozen features.}
\label{table:linear}
\begin{tabular}{llcc}
\toprule
\textbf{Method} & \textbf{Arch.} & \textbf{Epochs} & \textbf{Top-1} \\
\midrule
\multicolumn{4}{l}{\textit{Methods without view data augmentations}} \\
data2vec~\citep{baevski2022data2vec} & ViT-L/16 & 1600 & 77.3 \\
[1ex]
\multirow{3}{*}{MAE~\citep{he2021mae}}
 & ViT-B/16 & 1600 & 68.0 \\
 & ViT-L/16 & 1600 & 76.0 \\
 & ViT-H/14 & 1600 & 77.2 \\
 [1ex]
\multirow{3}{*}{I-JEPA~\citep{assran2023ijepa}}
 & ViT-B/16 & 600 & 72.9 \\
 & ViT-L/16 & 600 & 77.5 \\
 & ViT-H/14 & 300 & 79.3 \\
 [1ex]
\multirow{2}{*}{\model (ours)}
 & \cellcolor{pearDarker!12}ViT-B/16 & \cellcolor{pearDarker!12}600 & \cellcolor{pearDarker!12}73.5 \\
 & \cellcolor{pearDarker!12}ViT-L/16 & \cellcolor{pearDarker!12}600 & \cellcolor{pearDarker!12}78.3 \\
\midrule
\multicolumn{4}{l}{\textit{Methods using extra view data augmentations}} \\
\multirow{2}{*}{LeJEPA~\citep{balestriero2025lejepa}}
 & ViT-L/16 & 100 & 77.1 \\
 & ViT-H/14 & 100 & 79.0 \\
 [1ex]
iBOT~\citep{zhou2021ibot} & ViT-B/16 & 400 & 79.5 \\
[1ex]
\multirow{2}{*}{DINO~\citep{caron2021dino}}
 & ViT-B/16 & 100 & 78.2 \\
 & ViT-B/8 & 300 & 80.1 \\
 [1ex]
MSN~\citep{assran2022msn} & ViT-L/7 & 300 & 80.7 \\
 [1ex]
SimCLR v2~\citep{chen2020simclrv2} & RN152 ($2\times$) & 800 & 79.1 \\
\bottomrule
\end{tabular}
\end{table}
\begin{table}[!t]
\centering
\caption{{\bf ImageNet low-shot evaluation.} Top-1 accuracy with 1\% of ImageNet-1K labels, using full fine-tuning.}
\label{table:lowshot}
\setlength{\tabcolsep}{1.7mm}
\begin{tabular}{llcc}
\toprule
\textbf{Method} & \textbf{Arch.} & \textbf{Epochs} & \textbf{Top-1} \\
\midrule
\multicolumn{4}{l}{\textit{Methods without view data augmentations}} \\
data2vec~\citep{baevski2022data2vec} & ViT-L/16 & 1600 & 73.3 \\
[1ex]
\multirow{2}{*}{MAE~\citep{he2021mae}}
 & ViT-L/16 & 1600 & 67.1 \\
 & ViT-H/14 & 1600 & 71.5 \\
 [1ex]
\multirow{3}{*}{I-JEPA~\citep{assran2023ijepa}}
 & ViT-B/16$^\ddagger$ & 600 & 68.9 \\
 & ViT-L/16 & 600 & 69.4 \\
 & ViT-H/14 & 300 & 73.3 \\
 [1ex]
\multirow{2}{*}{\model (ours)}
 & \cellcolor{pearDarker!12}ViT-B/16 & \cellcolor{pearDarker!12}600 & \cellcolor{pearDarker!12}70.1 \\
 & \cellcolor{pearDarker!12}ViT-L/16 & \cellcolor{pearDarker!12}600 & \cellcolor{pearDarker!12}70.8 \\
\midrule
\multicolumn{4}{l}{\textit{Methods using extra view data augmentations}} \\
iBOT~\citep{zhou2021ibot} & ViT-B/16 & 400 & 69.7 \\
DINO~\citep{caron2021dino} & ViT-S/16 & 100 & 60.3 \\
SimCLR v2~\citep{chen2020simclrv2} & RN152 ($2\times$) & 800 & 70.2 \\
\bottomrule
\multicolumn{4}{l}{\footnotesize $^\ddagger$Our reproduction under the same protocol.} \\
\end{tabular}
\end{table}
\begin{table}[!t]
\centering
\caption{{\bf Transfer classification.} Top-1 linear probing accuracy on CIFAR-100, Places205, and iNaturalist 2018.}
\label{table:transfer_high}
\setlength{\tabcolsep}{1mm}
\begin{tabular}{llccc}
\toprule
\multicolumn{1}{l}{\textbf{Method}} & \textbf{Arch.} & \textbf{Cifar100} & \textbf{Places205} & \textbf{iNat18} \\
\midrule
\multicolumn{5}{l}{\textit{Methods without view data augmentations}} \\
data2vec~\citep{baevski2022data2vec} & ViT-L/16 & 59.6 & 36.9 & 10.9 \\
MAE~\citep{he2021mae} & ViT-H/14 & 77.3 & 55.0 & 32.9 \\
[1ex]
\multirow{3}{*}{I-JEPA~\citep{assran2023ijepa}}
 & ViT-B/16$^\dagger$ & 69.2 & 53.4 & 43.4 \\
 & ViT-L/16$^\dagger$ & 83.6 & 56.5 & 48.4 \\
 & ViT-H/14 & 87.5 & 58.4 & 47.6 \\
 [1ex]
\multirow{2}{*}{\model (ours)}
 & \cellcolor{pearDarker!12}ViT-B/16 & \cellcolor{pearDarker!12}70.0 & \cellcolor{pearDarker!12}54.3 & \cellcolor{pearDarker!12}43.2 \\
 & \cellcolor{pearDarker!12}ViT-L/16 & \cellcolor{pearDarker!12}85.2 & \cellcolor{pearDarker!12}57.1 & \cellcolor{pearDarker!12}44.3 \\
\midrule
\multicolumn{5}{l}{\textit{Methods using extra view data augmentations}} \\
DINO~\citep{caron2021dino} & ViT-B/8 & 84.9 & 57.9 & 55.9 \\
iBOT~\citep{zhou2021ibot} & ViT-L/16 & 88.3 & 60.4 & 57.3 \\
\bottomrule
\multicolumn{5}{l}{\footnotesize $^\dagger$From~\citep{bar2024stochastic}, under the same evaluation protocol.} \\
\end{tabular}
\end{table}
\begin{table}[!t]
\centering
\caption{{\bf Transfer low-level tasks.} Accuracy on object
counting (Clevr/Count) and depth prediction (Clevr/Dist).}
\label{table:clevr}
\setlength{\tabcolsep}{1.7mm}
\begin{tabular}{llcc}
\toprule
\multicolumn{1}{l}{\textbf{Method}} & \textbf{Arch.} & \textbf{Clevr/Count} & \textbf{Clevr/Dist} \\
\midrule
\multicolumn{4}{l}{\textit{Methods without view data augmentations}} \\
data2vec~\citep{baevski2022data2vec} & ViT-L/16 & 72.7 & 53.0 \\
MAE~\citep{he2021mae} & ViT-H/14 & 90.5 & 72.4 \\
[1ex]
\multirow{3}{*}{I-JEPA~\citep{assran2023ijepa}}
 & ViT-B/16$^\dagger$ & 82.2 & 70.7 \\
 & ViT-L/16$^\dagger$ & 85.6 & 71.2 \\
 & ViT-H/14 & 86.7 & 72.4 \\
 [1ex]
\multirow{2}{*}{\model (ours)}
 & \cellcolor{pearDarker!12}ViT-B/16 & \cellcolor{pearDarker!12}83.1 & \cellcolor{pearDarker!12}71.6 \\
 & \cellcolor{pearDarker!12}ViT-L/16 & \cellcolor{pearDarker!12}89.4 & \cellcolor{pearDarker!12}73.8 \\
\midrule
\multicolumn{4}{l}{\textit{Methods using extra view data augmentations}} \\
DINO~\citep{caron2021dino} & ViT-B/8 & 86.6 & 53.4 \\
iBOT~\citep{zhou2021ibot} & ViT-L/16 & 85.7 & 62.8 \\
\bottomrule
\end{tabular}
\end{table}

To assess the learned features for high-level semantic recognition without hand-crafted data augmentations, we conduct a series of benchmarks on ImageNet-1K~\citep{deng2009imagenet}. 

\paragraph{\bf ImageNet linear evaluation.}
We assess frozen feature quality by training a linear classifier on global average-pooled encoder outputs following~\citet{assran2023ijepa}. Results in \Cref{table:linear} show that \model surpasses I-JEPA at both backbone scales and exceeds reconstruction-based baselines~\citep{he2021mae,baevski2022data2vec} trained for substantially longer. At matched architecture it also exceeds LeJEPA~\citep{balestriero2025lejepa} without multi-view augmentations, at ${\approx}2.4\times$ its total encoder tokens (Appendix~\ref{subsec:disc-lejepa}).

\paragraph{\bf ImageNet low-shot evaluation.}
We evaluate label efficiency by fine-tuning the full pretrained model on 1\% of ImageNet-1K labels ($\approx$12 to 13 images per class), following the low-shot fine-tuning protocol of~\citet{assran2023ijepa}, which adapts the setup of~\citet{he2021mae}. Results in \Cref{table:lowshot} show that \model improves over I-JEPA by more than a point at both backbone scales, and exceeds methods relying on multi-crop augmentation~\citep{zhou2021ibot}. Modelling the distribution of latent completions is particularly valuable in this regime, where scarcity of labels places a premium on the richness of pretrained features.

\paragraph{\bf Transfer learning.}
To assess feature universality beyond ImageNet, we perform linear-probe transfer to CIFAR-100~\citep{krizhevsky2009learning}, Places205~\citep{zhou2014learning}, and iNaturalist 2018~\citep{van2018inaturalist}. These datasets span coarse object categories, scene recognition, and fine-grained species classification. Results are in \Cref{table:transfer_high}. \model improves over I-JEPA on object and scene recognition at both scales, while I-JEPA leads on iNaturalist 2018. With a ViT-L/16 backbone it surpasses MAE~\citep{he2021mae} and data2vec~\citep{baevski2022data2vec} and approaches I-JEPA's larger ViT-H/14 despite being smaller, narrowing the gap with the best view-invariance methods~\citep{caron2021dino,zhou2021ibot}. 

\subsection{Low-Level Vision Tasks}
\label{subsec:lowlevel}

We evaluate the encoder's ability to capture fine-grained spatial and structural information via linear-probe transfer to object counting and depth prediction on CLEVR~\citep{johnson2017clevr}. These tasks require preserving low-level geometric cues such as object multiplicity and relative distance, which methods enforcing view-invariance typically discard. Results are in \Cref{table:clevr}. \model improves over I-JEPA on both tasks at both backbone scales. This suggests that modelling the distribution of masked-region completions encourages the encoder to retain geometric structure a point-estimate predictor would suppress.

\subsection{Encoder Visualisation}
\label{subsec:encoder-viz}

\begin{figure*}[ht]
\centering
\setlength{\tabcolsep}{1.0pt}
\renewcommand{\arraystretch}{1}
\small
\begin{tabular}{ccccccc}
\multirow{2}{*}{Image} & \multicolumn{3}{c}{PCA} & \multicolumn{3}{c}{Attention} \\
\cmidrule(lr){2-4} \cmidrule(lr){5-7}
 & MAE & I-JEPA & S-JEPA & MAE & I-JEPA & S-JEPA \\
[2pt]
\includegraphics[width=0.13\linewidth]{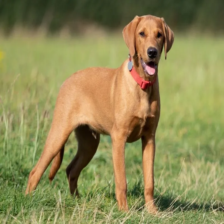} &
\includegraphics[width=0.13\linewidth]{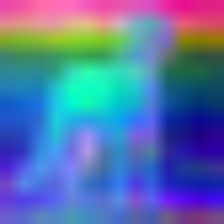} &
\includegraphics[width=0.13\linewidth]{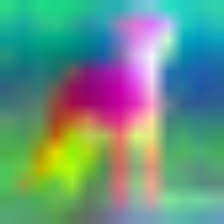} &
\includegraphics[width=0.13\linewidth]{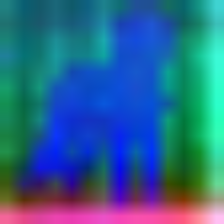} &
\includegraphics[width=0.13\linewidth]{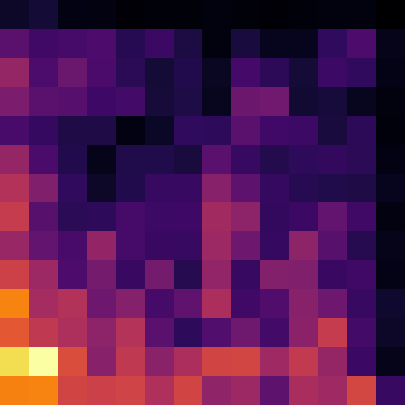} &
\includegraphics[width=0.13\linewidth]{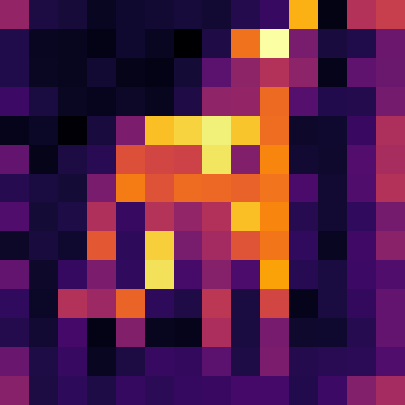} &
\includegraphics[width=0.13\linewidth]{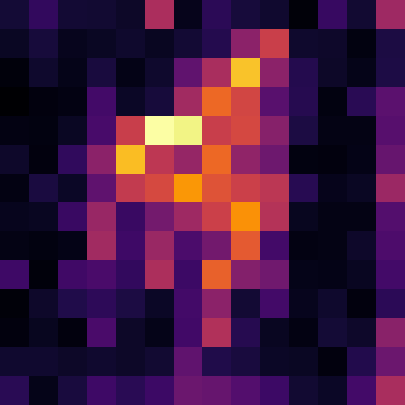} \\
[2pt]
\includegraphics[width=0.13\linewidth]{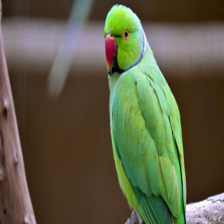} &
\includegraphics[width=0.13\linewidth]{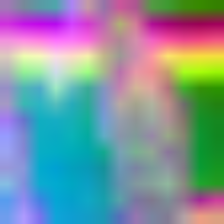} &
\includegraphics[width=0.13\linewidth]{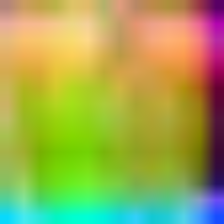} &
\includegraphics[width=0.13\linewidth]{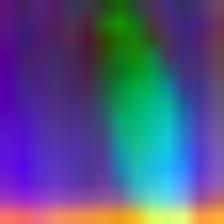} &
\includegraphics[width=0.13\linewidth]{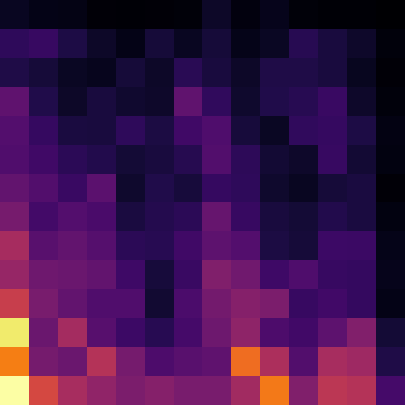} &
\includegraphics[width=0.13\linewidth]{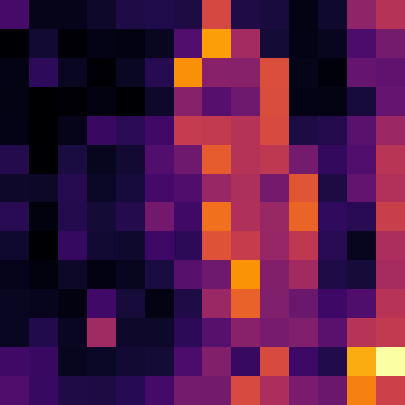} &
\includegraphics[width=0.13\linewidth]{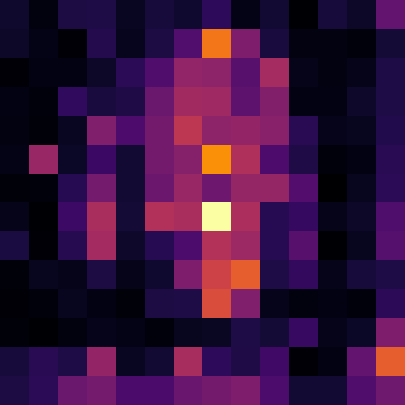} \\
[2pt]
\includegraphics[width=0.13\linewidth]{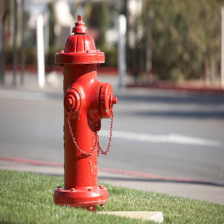} &
\includegraphics[width=0.13\linewidth]{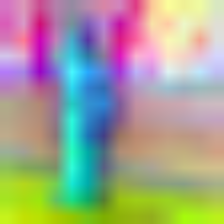} &
\includegraphics[width=0.13\linewidth]{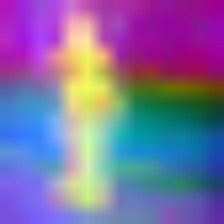} &
\includegraphics[width=0.13\linewidth]{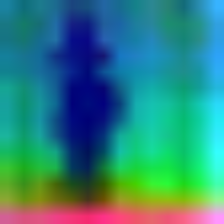} &
\includegraphics[width=0.13\linewidth]{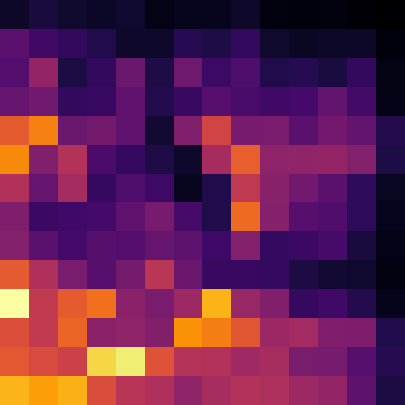} &
\includegraphics[width=0.13\linewidth]{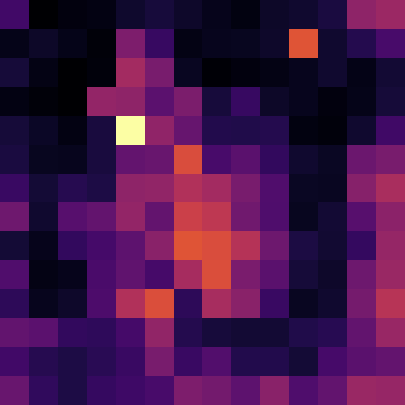} &
\includegraphics[width=0.13\linewidth]{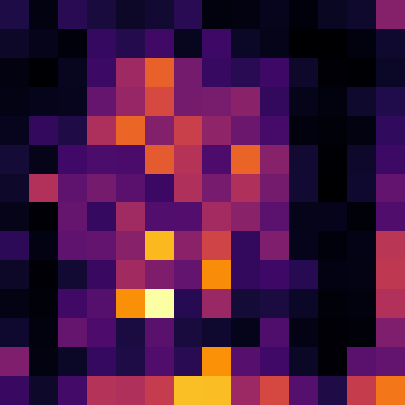} \\
\end{tabular}
\caption{{\bf PCA and attention visualisations of encoder features.} Each row shows the original image, the PCA projection (first three principal components of block-4 patch representations, foreground-masked, mapped to RGB), and the self-attention map (block 8, head 11, averaged over all query positions), across three ViT-B/16 encoders trained with different objectives (MAE~\cite{he2021mae}, I-JEPA~\cite{assran2023ijepa}, and \model). All methods recover coherent object/background structure in the PCA projection; the predictive objectives (I-JEPA, \model) additionally show attention concentrated on the primary object, in contrast to the more diffuse attention of the pixel-reconstruction objective (MAE). Neither map is supervised by any localisation signal: both emerge purely from the pretraining objective. Additional examples with more baselines are in Appendix~\Cref{fig:pca_vitb_appendix,fig:att_vitb_appendix}.}
\label{fig:pca}
\end{figure*}

A hallmark of successful self-supervised learning is the emergence of semantically meaningful structure without supervision~\citep{caron2021dino}. We apply PCA to patch tokens and project the first three principal components onto RGB channels, and separately visualise self-attention averaged over all query positions (no \texttt{[cls]} token), for three ViT-B/16 encoders trained with different objectives: MAE~\citep{he2021mae}, I-JEPA~\citep{assran2023ijepa}, and \model. As shown in \Cref{fig:pca}, all three objectives recover coherent object and background structure in the PCA projection, with maps that align with object boundaries and part-level structure. The two predictive objectives differ from the reconstruction objective in where they attend: I-JEPA and \model concentrate attention on the primary object and on fine detail without any localisation supervision, analogous to the emergent foreground segmentation in DINO~\citep{caron2021dino}, whereas the attention of the pixel-reconstruction objective is markedly more diffuse.

\subsection{Predictor Visualisation}
\label{subsec:predictor-viz}

We visualise two complementary aspects of the stochastic predictor: the diversity of pixel-space completions across independent latent draws, and the spatial structure of predictive uncertainty at inference.

\begin{figure*}[ht]
\centering
\setlength{\tabcolsep}{1.0pt}
\renewcommand{\arraystretch}{1}
\small
\begin{tabular}{ccccccc}
Image & Context & Sample 1 & Sample 2 & Sample 3 & Sample 4 & Sample 5 \\
[2pt]
\includegraphics[width=0.13\linewidth]{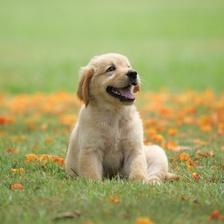} &
\includegraphics[width=0.13\linewidth]{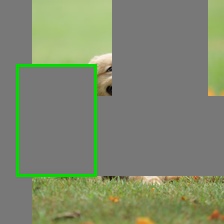} &
\includegraphics[width=0.13\linewidth]{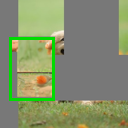} &
\includegraphics[width=0.13\linewidth]{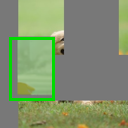} &
\includegraphics[width=0.13\linewidth]{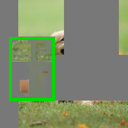} &
\includegraphics[width=0.13\linewidth]{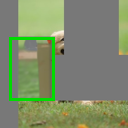} &
\includegraphics[width=0.13\linewidth]{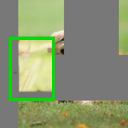} \\
[2pt]
\includegraphics[width=0.13\linewidth]{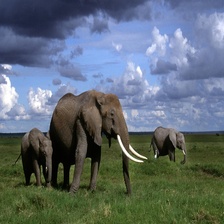} &
\includegraphics[width=0.13\linewidth]{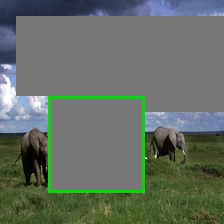} &
\includegraphics[width=0.13\linewidth]{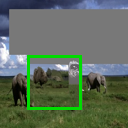} &
\includegraphics[width=0.13\linewidth]{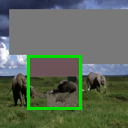} &
\includegraphics[width=0.13\linewidth]{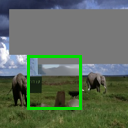} &
\includegraphics[width=0.13\linewidth]{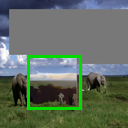} &
\includegraphics[width=0.13\linewidth]{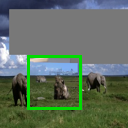} \\
[2pt]
\includegraphics[width=0.13\linewidth]{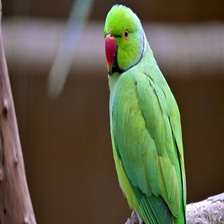} &
\includegraphics[width=0.13\linewidth]{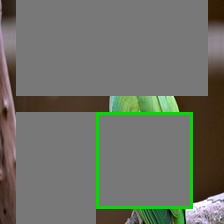} &
\includegraphics[width=0.13\linewidth]{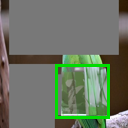} &
\includegraphics[width=0.13\linewidth]{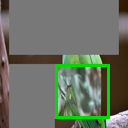} &
\includegraphics[width=0.13\linewidth]{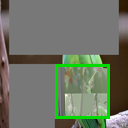} &
\includegraphics[width=0.13\linewidth]{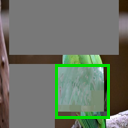} &
\includegraphics[width=0.13\linewidth]{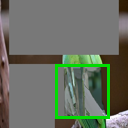} \\
\end{tabular}
\caption{{\bf Stochastic predictor completions via RCDM.} Each row shows a masked input and five completions, each from a distinct draw $\bz_i \sim \prior$ decoded via a lightweight diffusion decoder~\citep{bordes2022high} trained on frozen \model predictor representations. Samples vary in object identity and configuration, reflecting genuine uncertainty over the masked region encoded in $\bz$. Decoding uses a fixed noise seed across all five draws, so pixel-level variation reflects only the sampled latent and not decoder-side stochasticity.}
\label{fig:rcdm}
\end{figure*}

\paragraph{\bf Stochastic completions.}
We adapt the RCDM protocol of~\citet{assran2023ijepa}: a lightweight diffusion decoder~\citep{bordes2022high} trained on frozen \model predictor representations maps outputs back to pixel space. Five draws $\bz_i \sim \prior$ each yield a distinct predictor output decoded with a fixed noise seed, so variation originates entirely from the latent code. As shown in \Cref{fig:rcdm}, completions of structurally ambiguous regions vary in object identity and local configuration, which a deterministic predictor cannot produce by construction. Mean pairwise LPIPS~\citep{zhang2018lpips} between decoded draws is roughly four times higher over ambiguous regions than over visually determined ones such as sky or grass, confirming that $\bz$ allocates variability where the visible context genuinely under-determines the target rather than emitting uniform noise. This is consistent with context-dependent predictive diversity: genuinely ambiguous content yields diverse, semantically coherent completions, while visually determined content collapses to a single plausible outcome regardless of which $\bz$ is drawn.

\begin{figure*}[t]
\centering
\setlength{\tabcolsep}{1pt}
\renewcommand{\arraystretch}{1}
\small
\begin{tabular}{cccc@{\hspace{4pt}}cccc}
Image & I-JEPA $U_p$ & \model{} $U_p$ & \model{} error &
Image & I-JEPA $U_p$ & \model{} $U_p$ & \model{} error \\[2pt]
\includegraphics[width=0.12\linewidth]{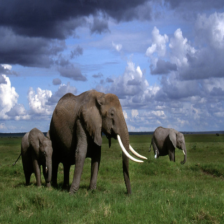} &
\includegraphics[width=0.12\linewidth]{figures/uncertainty/1/ijepa_Up_overlay.png} &
\includegraphics[width=0.12\linewidth]{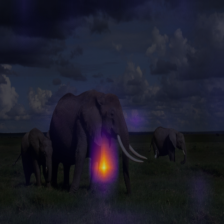} &
\includegraphics[width=0.12\linewidth]{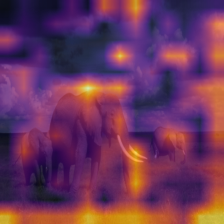} &
\includegraphics[width=0.12\linewidth]{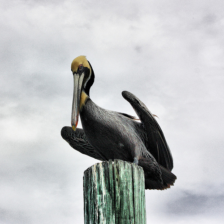} &
\includegraphics[width=0.12\linewidth]{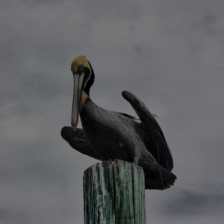} &
\includegraphics[width=0.12\linewidth]{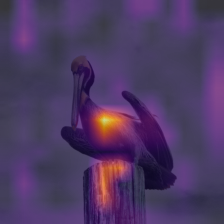} &
\includegraphics[width=0.12\linewidth]{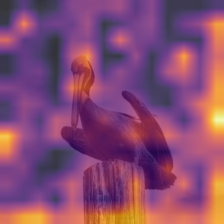} \\
[2pt]
\includegraphics[width=0.12\linewidth]{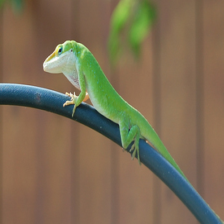} &
\includegraphics[width=0.12\linewidth]{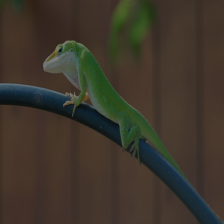} &
\includegraphics[width=0.12\linewidth]{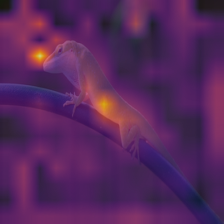} &
\includegraphics[width=0.12\linewidth]{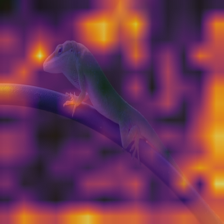} &
\includegraphics[width=0.12\linewidth]{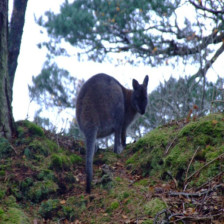} &
\includegraphics[width=0.12\linewidth]{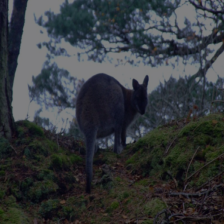} &
\includegraphics[width=0.12\linewidth]{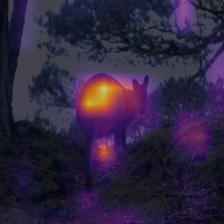} &
\includegraphics[width=0.12\linewidth]{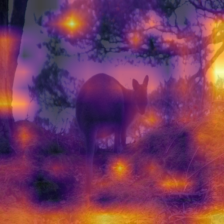} \\
[2pt]
\includegraphics[width=0.12\linewidth]{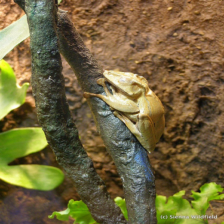} &
\includegraphics[width=0.12\linewidth]{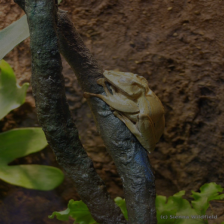} &
\includegraphics[width=0.12\linewidth]{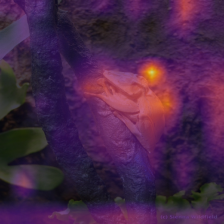} &
\includegraphics[width=0.12\linewidth]{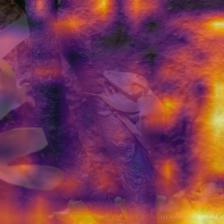} &
\includegraphics[width=0.12\linewidth]{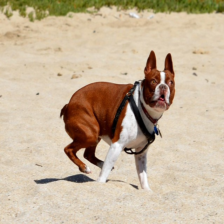} &
\includegraphics[width=0.12\linewidth]{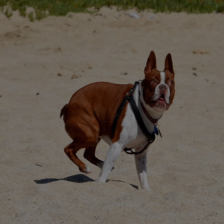} &
\includegraphics[width=0.12\linewidth]{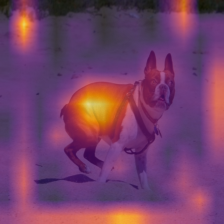} &
\includegraphics[width=0.12\linewidth]{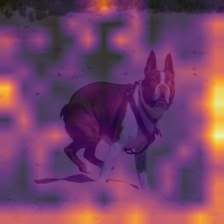}
\end{tabular}
\caption{{\bf Uncertainty and error maps.} $U_p$: predictive variance (Eq.~\eqref{eq:uncertainty}, $N{=}50$, full image as context) for I-JEPA, whose deterministic predictor gives $U_p{=}0$ everywhere, and \model{} (ViT-B/16). Error: squared distance of the mean \model{} prediction to the target, each $5{\times}5$-patch block masked in turn. The I-JEPA column carries no information about the underlying image, since its flat colour reflects only the normalisation; \model's $U_p$, by contrast, consistently concentrates on the animal or object rather than the background.}
\label{fig:uncertainty}
\end{figure*}

\paragraph{\bf Patch-level uncertainty maps.}
For each patch position $p$, we draw $N{=}50$ prior samples $\bz_1,\ldots,\bz_N \sim \prior$ and compute per-patch predictive variance:
\begin{equation}
    U_p = \frac{1}{d_y}\sum_{j=1}^{d_y} \mathrm{Var}_{i}\!\left[\hat{s}_{y,ij}^{(p)}\right].
    \label{eq:uncertainty}
\end{equation}

\noindent Values are reshaped, bilinearly upsampled to image resolution, and normalised per image; for I-JEPA, $U_p{=}0$ everywhere, shown in \Cref{fig:uncertainty} as a flat map. We compare \model's $U_p$ with prediction error, computed by masking each $5{\times}5$-patch block in turn and measuring the squared distance between the mean prediction and its target. \model places uncertainty around occlusions, object interiors, and semantic boundaries, with lower variance over uniform regions (\Cref{fig:uncertainty}). This suggests that the stochastic predictor captures meaningful, spatially structured ambiguity rather than simply injecting uniform noise. Under the pretraining masks (Appendix~\ref{subsec:app-uncertainty}), $U_p$ correlates with error across pooled patches (Spearman $\rho{=}0.33$, ViT-B/16), but only weakly within images (mean $\rho{=}0.16$, positive in 76\% of images). Thus, $U_p$ indicates prediction difficulty but does not reliably localise error. Raw predictions are underdispersed, potentially because $U_p$ excludes the observation-noise term $\sigma^2$ of \Cref{eq:gaussian-likelihood}; scalar recalibration reduces $\mathrm{ECE}_{\mathrm{reg}}$ from $0.18$ to $0.07$ (Appendix~\ref{subsec:disc-uncertainty}). The maps require only $N$ predictor passes after one encoder pass, with no labels or extra training.

\subsection{Cost of Stochasticity}
\label{subsec:cost}

At inference \model evaluates only the context encoder, the prior $\prior$, and the predictor; the posterior $\post$ is used during training and discarded thereafter, so the sole addition over I-JEPA is a 2-layer MLP producing the parameters of a 128-dimensional diagonal Gaussian. This amounts to a $1.01\times$ parameter increase, with FLOPs and single-pass wall-clock latency both within measurement noise of the baseline (ViT-B/16). The capabilities of \Cref{subsec:predictor-viz} are inexpensive but not free: uncertainty maps need $N$ predictor passes per encoder pass, and completions a separately trained diffusion decoder, neither requiring labels or backbone retraining.
\section{Conclusion}
\label{sec:conclusion}

We presented \model, which realises the auxiliary variable $\bz$ in JEPAs as a stochastic latent over the content of the masked region, with a learned context-conditioned prior and no reconstruction. A CVAE operating purely in representation space replaces I-JEPA's deterministic predictor with a probabilistic model whose objective recovers the I-JEPA loss as a deterministic special case. Under matched architectures and schedules, \model improves over I-JEPA on linear probing, low-shot, and most transfer tasks, exceeds LeJEPA at matched architecture on a longer schedule, produces error-correlated uncertainty maps at negligible cost, and generates diverse completions. Two limitations remain. $\bz$ is a single per-image latent rather than a per-target-region one and our largest backbone is ViT-L/16.

\section*{Acknowledgements}

We acknowledge the computational resources and support provided by the Imperial College Research Computing Service (\url{http://doi.org/10.14469/hpc/2232}), which enabled our experiments.

{
    \small
    \bibliographystyle{ieeenat_fullname}
    \bibliography{main}
}

\onecolumn
\appendix
\renewcommand{\thesubsection}{\Alph{section}.\arabic{subsection}}
\renewcommand{\thesubsubsection}{\Alph{section}.\arabic{subsection}.\arabic{subsubsection}}
\renewcommand{\thefigure}{\Alph{section}.\arabic{figure}}
\renewcommand{\thetable}{\Alph{section}.\arabic{table}}
\renewcommand{\theequation}{\Alph{section}.\arabic{equation}}
\counterwithin*{figure}{section}
\counterwithin*{table}{section}
\counterwithin*{equation}{section}
\setlength{\parindent}{0pt}
\setlength{\parskip}{0.5\baselineskip plus 2pt minus 1pt}
\setlength{\parfillskip}{0pt plus 1fil}
\makeatletter
\def\@listI{\leftmargin\leftmargini \topsep\z@ \parsep\parskip \itemsep\z@}\let\@listi\@listI \@listi
\def\@listii{\leftmargin\leftmarginii \labelwidth\leftmarginii\advance\labelwidth-\labelsep \topsep\z@ \parsep\parskip \itemsep\z@}
\def\@listiii{\leftmargin\leftmarginiii \labelwidth\leftmarginiii\advance\labelwidth-\labelsep \topsep\z@ \parsep\parskip \itemsep\z@}
\partopsep=\z@
\patchcmd\@startsection{\addvspace\@tempskipa}{\advance\@tempskipa-\parskip\addvspace\@tempskipa}{}{}
\patchcmd\@xsect{\vskip\@tempskipa}{\advance\@tempskipa-\parskip\vskip\@tempskipa}{}{}
\makeatother
\begin{center}
  \Large\textbf{\thetitle}\\[0.5em]
  Supplementary Material\\[1em]
\end{center}

\section{Proofs}
\label{sec:app-proofs}

We collect here the full proofs of all formal statements made in Section~\ref{sec:method}, together with supplementary remarks. Throughout, we write $\sx$, $\sy$, and $\bz$ for the context representation, target representation, and latent variable respectively, and use the shorthand $q_\psi \equiv \post$ and $p_\theta \equiv \prior$ where the conditioning is clear from context. The likelihood $p_\phi(\sy \mid \sx, \bz) = \mathcal{N}(\sy;\, g_\phi(\sx, \bz),\, \sigma^2 \mathbf{I})$ is parameterised by the predictor $g_\phi$, and $d_y$ denotes the dimensionality of the target representation $\sy$. As in Definition~\ref{def:sjepa}, the stop-gradient $\mathrm{sg}(\cdot)$ is applied to the target representation wherever it enters the prediction term; we suppress it in the derivations below, where it plays no role, and reinstate it explicitly whenever the \model objective itself is written out.

\subsection{Proof of Proposition 1: ELBO Lower Bound}
\label{subsec:proof-elbo}

\begin{proposition}{ELBO Lower Bound}{elbo}
Let $\prior$ be any prior over $\bz$ conditioned on $\sx$, let $\post$ be any approximate posterior, and let $g_\phi$ parameterise the likelihood $p_\phi(\sy \mid \sx, \bz)$. Then:
\begin{equation}
    \log p(\sy \mid \sx) \;\geq\; \underbrace{\mathbb{E}_{q_\psi}\!\left[\log p_\phi(\sy \mid \sx, \bz)\right] - \KL\!\left(\post \;\|\; \prior\right)}_{\mathrm{ELBO}(\phi, \psi, \theta;\, \sx, \sy)},
    \label{eq:app:elbo}
\end{equation}
with equality if and only if $\post = p(\bz \mid \sx, \sy)$ almost everywhere, where $p(\bz \mid \sx, \sy)$ denotes the true posterior under the generative model.
\end{proposition}

\begin{proof}
We begin from the marginal log-likelihood. By the law of total probability, the marginal likelihood of $\sy$ given $\sx$ is:
\begin{equation}
    p(\sy \mid \sx) = \int p_\phi(\sy \mid \sx, \bz)\, \prior\, d\bz.
\end{equation}
Taking the logarithm:
\begin{equation}
    \log p(\sy \mid \sx) = \log \int p_\phi(\sy \mid \sx, \bz)\, \prior\, d\bz.
\end{equation}
We now introduce the approximate posterior $q_\psi$ by multiplying and dividing inside the integral:
\begin{align}
    \log p(\sy \mid \sx)
    &= \log \int \frac{\post}{\post}\, p_\phi(\sy \mid \sx, \bz)\, \prior\, d\bz \\
    &= \log \int \post \cdot \frac{p_\phi(\sy \mid \sx, \bz)\, \prior}{\post}\, d\bz \\
    &= \log\, \mathbb{E}_{q_\psi}\!\left[\frac{p_\phi(\sy \mid \sx, \bz)\, \prior}{\post}\right].
\end{align}
Since $\log$ is concave, Jensen's inequality ($\log \mathbb{E}[X] \geq \mathbb{E}[\log X]$ for $X > 0$) gives:
\begin{equation}
    \log\, \mathbb{E}_{q_\psi}\!\left[\frac{p_\phi(\sy \mid \sx, \bz)\, \prior}{\post}\right]
    \;\geq\;
    \mathbb{E}_{q_\psi}\!\left[\log \frac{p_\phi(\sy \mid \sx, \bz)\, \prior}{\post}\right].
\end{equation}
Expanding the logarithm of the ratio:
\begin{align}
    \mathbb{E}_{q_\psi}\!\left[\log \frac{p_\phi(\sy \mid \sx, \bz)\, \prior}{\post}\right]
    &= \mathbb{E}_{q_\psi}\!\left[\log p_\phi(\sy \mid \sx, \bz)\right]
       + \mathbb{E}_{q_\psi}\!\left[\log \prior\right]
       - \mathbb{E}_{q_\psi}\!\left[\log \post\right] \\
    &= \mathbb{E}_{q_\psi}\!\left[\log p_\phi(\sy \mid \sx, \bz)\right]
       + \mathbb{E}_{q_\psi}\!\left[\log \frac{\prior}{\post}\right] \\
    &= \mathbb{E}_{q_\psi}\!\left[\log p_\phi(\sy \mid \sx, \bz)\right]
       - \mathbb{E}_{q_\psi}\!\left[\log \frac{\post}{\prior}\right] \\
    &= \mathbb{E}_{q_\psi}\!\left[\log p_\phi(\sy \mid \sx, \bz)\right]
       - \KL\!\left(q_\psi \;\|\; p_\theta\right),
\end{align}
where the last line uses the definition $\KL(q \| p) = \mathbb{E}_q[\log(q/p)]$. This establishes Eq.~\eqref{eq:app:elbo}.

\medskip
\noindent\textbf{Tightness condition.} Jensen's inequality is tight if and only if the random variable inside the expectation is constant $q_\psi$-almost everywhere:
\begin{equation}
    \frac{p_\phi(\sy \mid \sx, \bz)\, \prior}{\post} = c \quad q_\psi\text{-a.e.}
\end{equation}
for some constant $c > 0$. Integrating both sides over $\bz$ under $q_\psi$:
\begin{equation}
    \int \post \cdot c\, d\bz = c \cdot \underbrace{\int \post\, d\bz}_{=\,1} = c.
\end{equation}
On the other hand, integrating the numerator:
\begin{equation}
    \int p_\phi(\sy \mid \sx, \bz)\, \prior\, d\bz = p(\sy \mid \sx).
\end{equation}
Therefore $c = p(\sy \mid \sx)$, and rearranging the equality condition gives:
\begin{equation}
    \post = \frac{p_\phi(\sy \mid \sx, \bz)\, \prior}{p(\sy \mid \sx)}.
\end{equation}
By Bayes' theorem, the right-hand side is precisely the true posterior:
\begin{equation}
    p(\bz \mid \sx, \sy) = \frac{p_\phi(\sy \mid \sx, \bz)\, \prior}{p(\sy \mid \sx)}.
\end{equation}
Hence equality in the ELBO holds if and only if $\post = p(\bz \mid \sx, \sy)$ almost everywhere.
\end{proof}

\subsubsection{Remark: ELBO Gap Decomposition}
\label{subsec:remark-gap}

An alternative derivation makes the approximation gap explicit. Since $\log p(\sy \mid \sx)$ does not depend on $\bz$, we may write it as an expectation under any distribution over $\bz$:
\begin{equation}
    \log p(\sy \mid \sx) = \mathbb{E}_{q_\psi}\!\left[\log p(\sy \mid \sx)\right].
\end{equation}
By Bayes' theorem, $p(\sy \mid \sx) = p_\phi(\sy \mid \sx, \bz)\, \prior \,/\, p(\bz \mid \sx, \sy)$, so:
\begin{equation}
    \log p(\sy \mid \sx)
    = \mathbb{E}_{q_\psi}\!\left[\log \frac{p_\phi(\sy \mid \sx, \bz)\, \prior}{p(\bz \mid \sx, \sy)}\right].
\end{equation}
Multiplying and dividing inside the logarithm by $\post$:
\begin{align}
    \log p(\sy \mid \sx)
    &= \mathbb{E}_{q_\psi}\!\left[\log \frac{p_\phi(\sy \mid \sx, \bz)\, \prior}{\post} \cdot \frac{\post}{p(\bz \mid \sx, \sy)}\right] \\
    &= \mathbb{E}_{q_\psi}\!\left[\log \frac{p_\phi(\sy \mid \sx, \bz)\, \prior}{\post}\right]
       + \mathbb{E}_{q_\psi}\!\left[\log \frac{\post}{p(\bz \mid \sx, \sy)}\right] \\
    &= \underbrace{\mathbb{E}_{q_\psi}\!\left[\log p_\phi(\sy \mid \sx, \bz)\right] - \KL\!\left(q_\psi \;\|\; p_\theta\right)}_{\mathrm{ELBO}}
       + \underbrace{\KL\!\left(\post \;\|\; p(\bz \mid \sx, \sy)\right)}_{\geq\, 0}.
\end{align}
Since $\KL \geq 0$, we recover $\log p(\sy \mid \sx) \geq \mathrm{ELBO}$. The gap $\KL(q_\psi \| p(\bz \mid \sx, \sy))$ quantifies how far the learned posterior is from the true posterior; it vanishes (and the ELBO becomes tight) precisely when $q_\psi = p(\bz \mid \sx, \sy)$ almost everywhere.

\subsection{Proof of Proposition 2: Gaussian Likelihood Reduction}
\label{subsec:proof-gaussian}

\begin{proposition}{Gaussian Likelihood Reduction}{gaussian}
Let $p_\phi(\sy \mid \sx, \bz) = \mathcal{N}(\sy;\, g_\phi(\sx, \bz),\, \sigma^2 \mathbf{I}_{d_y})$ with $\sigma^2 > 0$ fixed. Then:
\begin{equation}
    -\mathbb{E}_{q_\psi}\!\left[\log p_\phi(\sy \mid \sx, \bz)\right]
    = \frac{1}{2\sigma^2}\,\mathbb{E}_{q_\psi}\!\left[\|\sy - g_\phi(\sx, \bz)\|^2\right]
    + \frac{d_y}{2}\log(2\pi\sigma^2),
    \label{eq:app:gaussian}
\end{equation}
and the negative ELBO equals, up to the additive constant $\frac{d_y}{2}\log(2\pi\sigma^2)$:
\begin{equation}
    -\mathrm{ELBO} = \frac{1}{2\sigma^2}\,\mathbb{E}_{q_\psi}\!\left[\|\sy - g_\phi(\sx, \bz)\|^2\right] + \KL\!\left(q_\psi \;\|\; p_\theta\right) + \frac{d_y}{2}\log(2\pi\sigma^2).
    \label{eq:app:neg-elbo}
\end{equation}
\end{proposition}

\begin{proof}
The probability density function of a $d_y$-dimensional isotropic Gaussian with mean $\bmu$ and variance $\sigma^2\mathbf{I}_{d_y}$ is:
\begin{equation}
    \mathcal{N}(\sy;\, \bmu,\, \sigma^2 \mathbf{I}_{d_y})
    = \frac{1}{(2\pi\sigma^2)^{d_y/2}} \exp\!\left(-\frac{\|\sy - \bmu\|^2}{2\sigma^2}\right).
\end{equation}
Taking the logarithm:
\begin{align}
    \log \mathcal{N}(\sy;\, \bmu,\, \sigma^2 \mathbf{I}_{d_y})
    &= \log\!\left(\frac{1}{(2\pi\sigma^2)^{d_y/2}}\right) - \frac{\|\sy - \bmu\|^2}{2\sigma^2} \\
    &= -\frac{d_y}{2}\log(2\pi) - \frac{d_y}{2}\log(\sigma^2) - \frac{\|\sy - \bmu\|^2}{2\sigma^2} \\
    &= -\frac{d_y}{2}\log(2\pi\sigma^2) - \frac{\|\sy - \bmu\|^2}{2\sigma^2}.
\end{align}
Setting $\bmu = g_\phi(\sx, \bz)$:
\begin{equation}
    \log p_\phi(\sy \mid \sx, \bz)
    = -\frac{d_y}{2}\log(2\pi\sigma^2) - \frac{\|\sy - g_\phi(\sx, \bz)\|^2}{2\sigma^2}.
\end{equation}
Negating:
\begin{equation}
    -\log p_\phi(\sy \mid \sx, \bz)
    = \frac{1}{2\sigma^2}\,\|\sy - g_\phi(\sx, \bz)\|^2 + \frac{d_y}{2}\log(2\pi\sigma^2).
\end{equation}
Taking the expectation under $q_\psi(\bz \mid \sx, \sy)$:
\begin{align}
    -\mathbb{E}_{q_\psi}\!\left[\log p_\phi(\sy \mid \sx, \bz)\right]
    &= \mathbb{E}_{q_\psi}\!\left[\frac{1}{2\sigma^2}\,\|\sy - g_\phi(\sx, \bz)\|^2 + \frac{d_y}{2}\log(2\pi\sigma^2)\right] \\
    &= \frac{1}{2\sigma^2}\,\mathbb{E}_{q_\psi}\!\left[\|\sy - g_\phi(\sx, \bz)\|^2\right] + \frac{d_y}{2}\log(2\pi\sigma^2),
\end{align}
where the constant passes through the expectation unchanged since it does not depend on $\bz$. This establishes Eq.~\eqref{eq:app:gaussian}.

Substituting into the negative ELBO:
\begin{align}
    -\mathrm{ELBO}
    &= -\mathbb{E}_{q_\psi}\!\left[\log p_\phi(\sy \mid \sx, \bz)\right] + \KL\!\left(q_\psi \;\|\; p_\theta\right) \\
    &= \frac{1}{2\sigma^2}\,\mathbb{E}_{q_\psi}\!\left[\|\sy - g_\phi(\sx, \bz)\|^2\right] + \KL\!\left(q_\psi \;\|\; p_\theta\right) + \frac{d_y}{2}\log(2\pi\sigma^2),
\end{align}
which yields Eq.~\eqref{eq:app:neg-elbo}.
\end{proof}

\subsubsection{Remark: Connection to \texorpdfstring{$\beta$}{beta}-VAE and the Role of \texorpdfstring{$\sigma^2$}{sigma squared}}
\label{subsec:remark-beta}

The negative ELBO in Eq.~\eqref{eq:app:neg-elbo} can be written as:
\begin{equation}
    -\mathrm{ELBO} = \frac{1}{2\sigma^2}\,\mathbb{E}_{q_\psi}\!\left[\|\sy - g_\phi(\sx, \bz)\|^2\right] + \KL\!\left(q_\psi \;\|\; p_\theta\right) + \mathrm{const}.
\end{equation}
Multiplying through by $2\sigma^2 > 0$ (which does not change the location of the optimum):
\begin{equation}
    2\sigma^2 \cdot (-\mathrm{ELBO})
    = \mathbb{E}_{q_\psi}\!\left[\|\sy - g_\phi(\sx, \bz)\|^2\right] + 2\sigma^2 \cdot \KL\!\left(q_\psi \;\|\; p_\theta\right) + \mathrm{const}.
\end{equation}
Comparing with the \model objective in Definition~\ref{def:sjepa}, in which the stop-gradient is applied to the target representation:
\begin{equation}
    \Ltotal = \underbrace{\mathbb{E}_{q_\psi}\!\left[\|\mathrm{sg}(\sy) - g_\phi(\sx, \bz)\|^2\right]}_{\Lpred} + \underbrace{\beta \cdot \KL\!\left(q_\psi \;\|\; p_\theta\right)}_{\Lreg},
\end{equation}
we see that the choice
\begin{equation}
    \beta = 2\sigma^2
\end{equation}
recovers exactly the scaled negative ELBO, corresponding to the exact variational lower bound under the Gaussian likelihood assumption. When $\beta$ is treated as a free hyperparameter independent of $\sigma^2$:
\begin{itemize}
    \item $\beta < 2\sigma^2$ weakens the KL regulariser relative to the exact bound, allowing $\bz$ to carry more target-specific information at the cost of a looser bound and potential posterior--prior mismatch at inference time.
    \item $\beta > 2\sigma^2$ strengthens the bottleneck beyond what the exact bound prescribes, forcing $\bz$ to encode only coarse-grained information and encouraging the predictor to rely more heavily on $\sx$ alone.
\end{itemize}
This is the same trade-off as in $\beta$-VAE~\citep{higgins2017betavae}, but operating entirely in representation space rather than observation space.

\subsection{Proof of Proposition 3: Closed-Form KL Divergence}
\label{subsec:proof-kl}

\begin{proposition}{Closed-Form KL Divergence}{kl}
Let $q_\psi = \mathcal{N}(\bmu_\psi, \mathrm{diag}(\bsig_\psi^2))$ and $p_\theta = \mathcal{N}(\bmu_\theta, \mathrm{diag}(\bsig_\theta^2))$ be two diagonal Gaussian distributions over $\mathbb{R}^{d_z}$. Then:
\begin{equation}
    \KL\!\left(q_\psi \;\|\; p_\theta\right)
    = \frac{1}{2} \sum_{k=1}^{d_z} \left(
        \log \frac{\sigma_{\theta,k}^2}{\sigma_{\psi,k}^2}
        + \frac{\sigma_{\psi,k}^2 + (\mu_{\psi,k} - \mu_{\theta,k})^2}{\sigma_{\theta,k}^2}
        - 1
    \right).
    \label{eq:app:kl}
\end{equation}
\end{proposition}

\begin{proof}
Since both distributions have diagonal covariance matrices, the KL divergence factorises into a sum of independent univariate KL divergences:
\begin{equation}
    \KL\!\left(q_\psi \;\|\; p_\theta\right) = \sum_{k=1}^{d_z} \KL\!\left(\mathcal{N}(\mu_{\psi,k}, \sigma_{\psi,k}^2) \;\|\; \mathcal{N}(\mu_{\theta,k}, \sigma_{\theta,k}^2)\right).
\end{equation}
It therefore suffices to derive the univariate case. Let $q = \mathcal{N}(\mu_q, \sigma_q^2)$ and $p = \mathcal{N}(\mu_p, \sigma_p^2)$. By definition:
\begin{equation}
    \KL(q \| p)
    = \int_{-\infty}^{+\infty} q(z) \log \frac{q(z)}{p(z)}\, dz
    = \mathbb{E}_q\!\left[\log q(z) - \log p(z)\right].
\end{equation}
The log-densities of the two univariate Gaussians are:
\begin{align}
    \log q(z) &= -\frac{1}{2}\log(2\pi) - \log \sigma_q - \frac{(z-\mu_q)^2}{2\sigma_q^2}, \\[4pt]
    \log p(z) &= -\frac{1}{2}\log(2\pi) - \log \sigma_p - \frac{(z-\mu_p)^2}{2\sigma_p^2}.
\end{align}
Taking the difference:
\begin{align}
    \log q(z) - \log p(z)
    &= -\log\sigma_q + \log\sigma_p - \frac{(z-\mu_q)^2}{2\sigma_q^2} + \frac{(z-\mu_p)^2}{2\sigma_p^2} \\
    &= \frac{1}{2}\log\frac{\sigma_p^2}{\sigma_q^2} - \frac{(z-\mu_q)^2}{2\sigma_q^2} + \frac{(z-\mu_p)^2}{2\sigma_p^2}.
\end{align}
Taking the expectation under $q$, and using the identities:
\begin{align}
    \mathbb{E}_q\!\left[(z - \mu_q)^2\right] &= \sigma_q^2, \\
    \mathbb{E}_q\!\left[(z - \mu_p)^2\right] &= \mathrm{Var}_q(z) + (\mathbb{E}_q[z] - \mu_p)^2 = \sigma_q^2 + (\mu_q - \mu_p)^2,
\end{align}
we obtain:
\begin{align}
    \KL(q \| p)
    &= \frac{1}{2}\log\frac{\sigma_p^2}{\sigma_q^2}
       - \frac{\sigma_q^2}{2\sigma_q^2}
       + \frac{\sigma_q^2 + (\mu_q - \mu_p)^2}{2\sigma_p^2} \\
    &= \frac{1}{2}\log\frac{\sigma_p^2}{\sigma_q^2}
       - \frac{1}{2}
       + \frac{\sigma_q^2 + (\mu_q - \mu_p)^2}{2\sigma_p^2} \\
    &= \frac{1}{2}\left(
        \log\frac{\sigma_p^2}{\sigma_q^2}
        + \frac{\sigma_q^2 + (\mu_q - \mu_p)^2}{\sigma_p^2}
        - 1
    \right).
\end{align}
Summing over $k = 1, \ldots, d_z$ and substituting $(\mu_q, \sigma_q) \leftarrow (\mu_{\psi,k}, \sigma_{\psi,k})$ and $(\mu_p, \sigma_p) \leftarrow (\mu_{\theta,k}, \sigma_{\theta,k})$:
\begin{equation}
    \KL\!\left(q_\psi \;\|\; p_\theta\right)
    = \frac{1}{2} \sum_{k=1}^{d_z} \left(
        \log \frac{\sigma_{\theta,k}^2}{\sigma_{\psi,k}^2}
        + \frac{\sigma_{\psi,k}^2 + (\mu_{\psi,k} - \mu_{\theta,k})^2}{\sigma_{\theta,k}^2}
        - 1
    \right),
\end{equation}
which is Eq.~\eqref{eq:app:kl}.
\end{proof}

\subsubsection{Remark: Special Case with Standard Normal Prior}
\label{subsec:remark-standard-normal-kl}

When the prior is the standard normal $p(\bz) = \mathcal{N}(\mathbf{0}, \mathbf{I})$, \ie, $\mu_{\theta,k} = 0$ and $\sigma_{\theta,k}^2 = 1$ for all $k$, Proposition~\ref{proposition:kl} simplifies. Each term in the sum becomes:
\begin{equation}
    \log \frac{1}{\sigma_{\psi,k}^2}
    + \frac{\sigma_{\psi,k}^2 + \mu_{\psi,k}^2}{1}
    - 1
    = -\log\sigma_{\psi,k}^2 + \sigma_{\psi,k}^2 + \mu_{\psi,k}^2 - 1,
\end{equation}
yielding the well-known identity~\citep{kingma2013vae}:
\begin{equation}
    \KL\!\left(q_\psi \;\|\; \mathcal{N}(\mathbf{0}, \mathbf{I})\right)
    = \frac{1}{2}\sum_{k=1}^{d_z}\left(
        \sigma_{\psi,k}^2 + \mu_{\psi,k}^2 - 1 - \log \sigma_{\psi,k}^2
    \right).
    \label{eq:app:kl-standard}
\end{equation}
Equivalently, writing $-\KL$ as it often appears in the VAE objective:
\begin{equation}
    -\KL\!\left(q_\psi \;\|\; \mathcal{N}(\mathbf{0}, \mathbf{I})\right)
    = \frac{1}{2}\sum_{k=1}^{d_z}\left(
        1 + \log \sigma_{\psi,k}^2 - \mu_{\psi,k}^2 - \sigma_{\psi,k}^2
    \right).
\end{equation}
In \model the prior is context-conditioned ($\prior$ with learned $\bmu_\theta(\sx)$ and $\bsig_\theta(\sx)$), so the full expression of Proposition~\ref{proposition:kl} is required. We include this special case for completeness and to highlight the connection to the standard VAE objective. We note finally where the Gaussian assumption enters. \Cref{proposition:elbo} holds for any prior and any approximate posterior, and \Cref{proposition:gaussian} requires only that the likelihood be isotropic Gaussian; the diagonal-Gaussian family is assumed only for the closed form of \Cref{proposition:kl} and for the reparameterisation used in practice, so richer families are substitutable without altering the bound (\Cref{subsec:disc-vamp}).

\subsection{Proof of Theorem 1: I-JEPA Recovery}
\label{subsec:proof-ijepa}

\begin{theorem}{I-JEPA Recovery}{ijepa}
Let $\bsig_\psi(\sx, \sy) \to \mathbf{0}$ so that the posterior $\post$ concentrates at $\bmu_\psi(\sx, \sy)$, and let $\bsig_\theta(\sx)$ remain bounded away from zero. Then setting $\beta = 0$ in Eq.~\eqref{eq:total-loss} recovers the I-JEPA objective of Definition~\ref{def:jepa} in functional form:
\begin{equation}
    \Ltotal\big|_{\beta=0,\, \bsig_\psi \to \mathbf{0}} = \|\mathrm{sg}(\sy) - g_\phi(\sx, \bmu_\psi)\|^2.
\end{equation}
\end{theorem}

\begin{proof}
We analyse each term in $\Ltotal = \Lpred + \beta \cdot \Lreg$ separately.

\medskip
\noindent\textbf{Step 1: Prediction term.} The prediction loss is:
\begin{equation}
    \Lpred = \mathbb{E}_{q_\psi}\!\left[\|\mathrm{sg}(\sy) - g_\phi(\sx, \bz)\|^2\right]
    = \int \post\, \|\mathrm{sg}(\sy) - g_\phi(\sx, \bz)\|^2\, d\bz.
\end{equation}
As $\bsig_\psi \to \mathbf{0}$, the Gaussian posterior $\post$ converges weakly to the Dirac measure $\delta_{\bmu_\psi}$ centred at $\bmu_\psi = \bmu_\psi(\sx, \sy)$. By the defining property of the Dirac delta:
\begin{equation}
    \int \delta_{\bmu_\psi}(\bz)\, \|\mathrm{sg}(\sy) - g_\phi(\sx, \bz)\|^2\, d\bz = \|\mathrm{sg}(\sy) - g_\phi(\sx, \bmu_\psi)\|^2.
\end{equation}
Therefore:
\begin{equation}
    \lim_{\bsig_\psi \to \mathbf{0}} \Lpred = \|\mathrm{sg}(\sy) - g_\phi(\sx, \bmu_\psi)\|^2.
\end{equation}

\medskip
\noindent\textbf{Step 2: KL term.} From Proposition~\ref{proposition:kl}, the KL divergence is:
\begin{equation}
    \KL(q_\psi \| p_\theta) = \frac{1}{2}\sum_{k=1}^{d_z}\left(
        \log\frac{\sigma_{\theta,k}^2}{\sigma_{\psi,k}^2}
        + \frac{\sigma_{\psi,k}^2 + (\mu_{\psi,k} - \mu_{\theta,k})^2}{\sigma_{\theta,k}^2}
        - 1
    \right).
\end{equation}
As $\sigma_{\psi,k} \to 0^+$ with $\sigma_{\theta,k}$ bounded away from zero:
\begin{equation}
    \log\frac{\sigma_{\theta,k}^2}{\sigma_{\psi,k}^2} \;\longrightarrow\; +\infty,
    \qquad
    \frac{\sigma_{\psi,k}^2}{\sigma_{\theta,k}^2} \;\longrightarrow\; 0,
    \qquad
    \frac{(\mu_{\psi,k} - \mu_{\theta,k})^2}{\sigma_{\theta,k}^2} \;\text{remains finite}.
\end{equation}
The dominant term is $\log(\sigma_{\theta,k}^2 / \sigma_{\psi,k}^2) \to +\infty$, so:
\begin{equation}
    \KL(q_\psi \| p_\theta) \;\longrightarrow\; +\infty.
\end{equation}
However, the loss multiplies this divergent quantity by $\beta$. The formal argument proceeds by first fixing $\beta = 0$, which zeroes out the regulariser for \emph{all} values of $\bsig_\psi$:
\begin{equation}
    \beta \cdot \KL(q_\psi \| p_\theta)\big|_{\beta=0} = 0 \cdot \KL(q_\psi \| p_\theta) = 0,
\end{equation}
and then taking $\bsig_\psi \to \mathbf{0}$ in the prediction term. The order of operations is: set $\beta = 0$ first, then take the limit $\bsig_\psi \to \mathbf{0}$.

\medskip
\noindent\textbf{Step 3: Combined.} Under the two conditions $\beta = 0$ and $\bsig_\psi \to \mathbf{0}$:
\begin{align}
    \Ltotal\big|_{\beta=0,\, \bsig_\psi \to \mathbf{0}}
    &= \lim_{\bsig_\psi \to \mathbf{0}} \Lpred + 0 \\
    &= \|\mathrm{sg}(\sy) - g_\phi(\sx, \bmu_\psi)\|^2.
\end{align}

\medskip
\noindent\textbf{Step 4: Equivalence to I-JEPA.} Since $\bmu_\psi(\sx, \sy)$ is a deterministic function of $(\sx, \sy)$, this is precisely the I-JEPA objective of Definition~\ref{def:jepa} with $\bz = \bmu_\psi(\sx, \sy)$ playing the role of the deterministic auxiliary variable. The stop-gradient on the target encoder output is retained exactly as in Definition~\ref{def:sjepa}, consistent with I-JEPA~\citep{assran2023ijepa}.

Note that in I-JEPA, the auxiliary variable $\bz$ is a shared learnable mask token with positional embeddings (independent of $\sy$), whereas in the deterministic limit of \model, $\bz = \bmu_\psi(\sx, \sy)$ is image-specific. The recovery is at the level of the \emph{loss functional form} $\|\sy - g(\sx, \bz)\|^2$, not the specific parameterisation of the auxiliary input.
\end{proof}

\subsection{Proof of Lemma 1: KL as Information Bottleneck}
\label{subsec:proof-ib}

\begin{lemma}{KL as Information Bottleneck}{ib}
The expected KL divergence in the \model objective upper-bounds the conditional mutual information between $\bz$ and $\sy$ given $\sx$ under $q_\psi$:
\begin{equation}
    I_{q_\psi}(\bz;\, \sy \mid \sx)
    \;\leq\;
    \mathbb{E}_{\sx, \sy}\!\left[\KL\!\left(\post \;\|\; \prior\right)\right].
    \label{eq:app:ib}
\end{equation}
Hence $\beta$ bounds how much information about $\sy$ is encoded in $\bz$ beyond what is already explained by $\sx$.
\end{lemma}

\begin{proof}
\noindent\textbf{Step 1: Definition of conditional mutual information.} Under the joint distribution induced by $q_\psi$, the conditional mutual information is defined as:
\begin{align}
    I_{q_\psi}(\bz;\, \sy \mid \sx)
    &= \mathbb{E}_{\sx, \sy}\!\left[\KL\!\left(\post \;\|\; q_\psi(\bz \mid \sx)\right)\right],
    \label{eq:app:cmi-def}
\end{align}
where the aggregate posterior is the marginal obtained by averaging the posterior over all targets given context:
\begin{equation}
    q_\psi(\bz \mid \sx) = \mathbb{E}_{\sy \mid \sx}\!\left[\post\right] = \int p(\sy \mid \sx)\, \post\, d\sy.
\end{equation}

\medskip
\noindent\textbf{Step 2: Variational upper bound.} For any reference distribution $r(\bz \mid \sx)$, we claim:
\begin{equation}
    \mathbb{E}_{\sy \mid \sx}\!\left[\KL\!\left(\post \;\|\; q_\psi(\bz \mid \sx)\right)\right]
    \;\leq\;
    \mathbb{E}_{\sy \mid \sx}\!\left[\KL\!\left(\post \;\|\; r(\bz \mid \sx)\right)\right].
    \label{eq:app:kl-mixture-bound}
\end{equation}
To see this, consider the difference for a fixed $(\sx, \sy)$:
\begin{align}
    &\KL\!\left(\post \;\|\; r(\bz \mid \sx)\right) - \KL\!\left(\post \;\|\; q_\psi(\bz \mid \sx)\right) \\
    &= \mathbb{E}_{q_\psi(\bz \mid \sx, \sy)}\!\left[\log \frac{\post}{r(\bz \mid \sx)}\right]
       - \mathbb{E}_{q_\psi(\bz \mid \sx, \sy)}\!\left[\log \frac{\post}{q_\psi(\bz \mid \sx)}\right] \\
    &= \mathbb{E}_{q_\psi(\bz \mid \sx, \sy)}\!\left[\log \frac{\post}{r(\bz \mid \sx)} - \log \frac{\post}{q_\psi(\bz \mid \sx)}\right] \\
    &= \mathbb{E}_{q_\psi(\bz \mid \sx, \sy)}\!\left[\log \frac{q_\psi(\bz \mid \sx)}{r(\bz \mid \sx)}\right].
\end{align}
Note that the integrand $\log(q_\psi(\bz \mid \sx) / r(\bz \mid \sx))$ does not depend on $\sy$. Taking the expectation over $\sy \mid \sx$:
\begin{align}
    &\mathbb{E}_{\sy \mid \sx}\!\left[\KL(\post \| r) - \KL(\post \| q_\psi(\bz \mid \sx))\right] \\
    &= \mathbb{E}_{\sy \mid \sx}\!\left[\mathbb{E}_{q_\psi(\bz \mid \sx, \sy)}\!\left[\log \frac{q_\psi(\bz \mid \sx)}{r(\bz \mid \sx)}\right]\right] \\
    &= \int \underbrace{\left(\int p(\sy \mid \sx)\, q_\psi(\bz \mid \sx, \sy)\, d\sy\right)}_{=\, q_\psi(\bz \mid \sx)} \log \frac{q_\psi(\bz \mid \sx)}{r(\bz \mid \sx)}\, d\bz \\
    &= \KL\!\left(q_\psi(\bz \mid \sx) \;\|\; r(\bz \mid \sx)\right) \\
    &\geq 0,
\end{align}
where the last inequality follows from the non-negativity of KL divergence. This establishes Eq.~\eqref{eq:app:kl-mixture-bound}.

\medskip
\noindent\textbf{Step 3: Choosing $r = p_\theta$.} Setting $r(\bz \mid \sx) = \prior$ and taking the outer expectation over $\sx$:
\begin{align}
    I_{q_\psi}(\bz;\, \sy \mid \sx)
    &= \mathbb{E}_{\sx, \sy}\!\left[\KL\!\left(\post \;\|\; q_\psi(\bz \mid \sx)\right)\right] \\
    &\leq \mathbb{E}_{\sx, \sy}\!\left[\KL\!\left(\post \;\|\; \prior\right)\right].
\end{align}

\medskip
\noindent\textbf{Step 4: Interpretation.} Since the regularisation term in the \model objective is $\Lreg = \beta \cdot \mathbb{E}[\KL(q_\psi \| p_\theta)]$, and this upper-bounds $\beta \cdot I_{q_\psi}(\bz; \sy \mid \sx)$, minimising $\Lreg$ constrains the target-specific information that $\bz$ can carry beyond what is predictable from the context $\sx$ via the prior. Larger $\beta$ imposes a tighter bottleneck, forcing $\bz$ to discard fine-grained target details; smaller $\beta$ allows $\bz$ to encode richer target-specific information at the cost of potential posterior--prior mismatch at test time.
\end{proof}

\subsection{Remark: Prior Learning via Aggregate Posterior Matching}
\label{subsec:remark-prior}

Since $\prior$ appears only in the KL term $\Lreg$, optimising $\Ltotal$ with respect to the prior parameters $\theta$ alone is equivalent to:
\begin{equation}
    \min_\theta\; \mathbb{E}_{\sx, \sy}\!\left[\KL\!\left(\post \;\|\; \prior\right)\right].
\end{equation}
We now show that the optimal prior, for a fixed posterior $q_\psi$, equals the aggregate posterior. For a given $\sx$, define the aggregate posterior:
\begin{equation}
    q_\psi^{\mathrm{agg}}(\bz \mid \sx) \;\triangleq\; \mathbb{E}_{\sy \mid \sx}\!\left[\post\right] = \int p(\sy \mid \sx)\, \post\, d\sy.
\end{equation}
For any candidate $p_\theta(\bz \mid \sx)$, we decompose the expected KL:
\begin{align}
    &\mathbb{E}_{\sy \mid \sx}\!\left[\KL\!\left(\post \;\|\; p_\theta(\bz \mid \sx)\right)\right] \\
    &= \mathbb{E}_{\sy \mid \sx}\!\left[\int \post \log \frac{\post}{p_\theta(\bz \mid \sx)}\, d\bz\right] \\
    &= \mathbb{E}_{\sy \mid \sx}\!\left[\int \post \log \frac{\post}{q_\psi^{\mathrm{agg}}(\bz \mid \sx)}\, d\bz\right]
       + \mathbb{E}_{\sy \mid \sx}\!\left[\int \post \log \frac{q_\psi^{\mathrm{agg}}(\bz \mid \sx)}{p_\theta(\bz \mid \sx)}\, d\bz\right] \\
    &= \mathbb{E}_{\sy \mid \sx}\!\left[\KL\!\left(\post \;\|\; q_\psi^{\mathrm{agg}}(\bz \mid \sx)\right)\right]
       + \int \underbrace{\mathbb{E}_{\sy \mid \sx}\!\left[\post\right]}_{=\, q_\psi^{\mathrm{agg}}(\bz \mid \sx)} \log \frac{q_\psi^{\mathrm{agg}}(\bz \mid \sx)}{p_\theta(\bz \mid \sx)}\, d\bz \\
    &= \underbrace{\mathbb{E}_{\sy \mid \sx}\!\left[\KL\!\left(\post \;\|\; q_\psi^{\mathrm{agg}}(\bz \mid \sx)\right)\right]}_{\text{does not depend on } \theta}
       + \underbrace{\KL\!\left(q_\psi^{\mathrm{agg}}(\bz \mid \sx) \;\|\; p_\theta(\bz \mid \sx)\right)}_{\geq\, 0}.
\end{align}
The first term is independent of $\theta$. The second term is non-negative and equals zero if and only if $p_\theta(\bz \mid \sx) = q_\psi^{\mathrm{agg}}(\bz \mid \sx)$ almost everywhere. Therefore the unique minimiser is:
\begin{equation}
    p_\theta^*(\bz \mid \sx) = q_\psi^{\mathrm{agg}}(\bz \mid \sx) = \mathbb{E}_{\sy \mid \sx}\!\left[\post\right].
\end{equation}

At convergence, the learned prior $\prior$ therefore approximates the full distribution over latent codes that arise when the target is observed, internalising the statistics of all plausible completions from the visible context. This is the mechanism by which $\bz$ sampled from the prior at inference time still carries meaningful information about the distribution of masked content.

\section{Implementation Details}
\label{sec:app-implementation}

\subsection{Pretraining}
\label{subsec:app-pretraining}

\paragraph{\bf Encoder.}
We use vision transformer~\citep{dosovitskiy2020vit} architectures for the context encoder, target encoder, and predictor, following~\citet{assran2023ijepa}. The context and target encoders correspond to standard ViT architectures: ViT-B/16 (embedding dimension 768, depth 12, 12 heads) and ViT-L/16 (embedding dimension 1024, depth 24, 16 heads), both operating at resolution $224{\times}224$. We do not use a \texttt{[cls]} token; the target encoder output is average-pooled to produce a global image representation. The target encoder weights are initialised identically to the context encoder and subsequently updated via exponential moving average~\citep{grill2020byol} with momentum linearly scheduled from 0.996 to 1.0 throughout pretraining~\citep{caron2021dino,assran2022msn}.

\paragraph{\bf Predictor.}
The predictor is a narrow vision transformer with embedding dimension 384, 12 attention heads, and MLP ratio 4, following the lightweight predictor design of~\citet{assran2023ijepa}; its depth is 6 for ViT-B/16 and 12 for ViT-L/16, matching the backbone-dependent predictor depth of~\citet{assran2023ijepa}. Context encoder tokens are linearly projected into the predictor dimension, and sine-cosine positional embeddings~\citep{vaswani2017attention} are added. For \model, $\bz \in \mathbb{R}^{128}$ is linearly projected to the predictor dimension and prepended as a single prefix token. Learnable mask tokens with sine-cosine positional embeddings are appended for each target position, and a final linear layer projects back to the encoder dimension. Both the prior $\prior$ and posterior $\post$ are 2-layer MLPs with GELU activations, mapping to the parameters $(\bmu, \log \bsig)$ of a diagonal Gaussian over $\mathbb{R}^{128}$.

\paragraph{\bf Masking.} 
We use the \texttt{multi-block} masking strategy of~\citet{assran2023ijepa} as the default for \model. We sample $4$ possibly overlapping target block masks with random scale in the range $(0.15, 0.2)$ and aspect ratio in the range $(0.75, 1.5)$. We sample $1$ context block mask with random scale in the range $(0.85, 1.0)$ and unit aspect ratio. We subsequently eliminate any regions in the context block mask that overlap with any of the 4 target block masks. The context block mask and target block masks are sampled independently for each image in the mini-batch. 

\paragraph{\bf Optimisation.}
We use AdamW~\citep{loshchilov2017adamw} with batch size of 2048 for ViT-B/16 and 1024 for ViT-L/16 on 4 A100 GPUs with \texttt{bfloat16} mixed precision. The learning rate is warmed up linearly from $2{\times}10^{-4}$ to $10^{-3}$ over 40 epochs, then decayed via cosine schedule to $10^{-6}$. Weight decay follows a cosine schedule from 0.04 to 0.4 \citep{caron2021dino,assran2022msn}. The KL weight is linearly warmed up from 0 to $\beta{=}0.010$ over the first 30\% of training, with no free-bits floor ($\lambda{=}0$). We train for 600 epochs on ImageNet-1K~\citep{deng2009imagenet}.

\subsection{Downstream Tasks}
\label{subsec:app-downstream}

\paragraph{\bf Linear evaluation.} 
We follow the linear evaluation protocol of~\citet{assran2023ijepa} that adapts the setup of \citep{he2021mae}. We freeze the encoder and use the average-pooled patch representation of the last layer. Following~\citet{he2021mae,assran2023ijepa}, we use the LARS~\citep{you2017lars} optimizer with batch size 4096 and train the linear classifier for 50 epochs with a learning rate decayed step-wise by a factor of 10 every 15 epochs, sweeping reference learning rates $\{0.001, 0.01, 0.05\}$ and weight decay values $\{0.0,\, 5{\times}10^{-4}\}$. No augmentations beyond random resized crop and horizontal flip are applied. For transfer to CIFAR-100~\citep{krizhevsky2009learning}, Places205~\citep{zhou2014learning}, iNaturalist 2018~\citep{van2018inaturalist}, CLEVR/Count and CLEVR/Dist~\citep{johnson2017clevr}, we use $\ell_2$-regularised logistic regression on frozen average-pooled features, following~\citet{assran2023ijepa}. For CLEVR tasks we use only center crop and horizontal flip to preserve object count and depth cues.

\paragraph{\bf Low-shot evaluation.} 
We follow the low-shot fine-tuning protocol of~\citet{assran2023ijepa}, adapting the fine-tuning setup of~\citet{he2021mae}. We fine-tune the full model for 50 epochs on the ImageNet-1K subset using AdamW with cosine learning rate decay, batch size 512, layer-wise learning rate decay of 0.75, and label smoothing of 0.1. We use RandAugment data augmentations but do not apply mixup, cutmix, random erasing, or drop path, following~\citet{assran2023ijepa}. We use learning rate $3{\times}10^{-5}$ for both backbones, with weight decay $5{\times}10^{-2}$ for ViT-B/16 and $4{\times}10^{-1}$ for ViT-L/16.

\subsection{Visualization}
\label{subsec:app-viz-protocols}

We describe the protocols used to visualise encoder features and assess the spatial structure and calibration of predictive uncertainty.

\paragraph{PCA.}
We extract block-4 patch features from ViT-B/16 encoders at $224{\times}224$ resolution without gradients and $\ell_2$-normalise each patch. A one-component PCA over the $14{\times}14$ patch grid is thresholded at its median to select foreground patches. We fit a three-component PCA on this subset and project all patches onto that basis. Each channel is clipped at the 1st and 99th percentiles, rescaled to $[0,1]$, and bilinearly upsampled to image resolution to form an RGB map.

\paragraph{Attention.}
We extract self-attention from block 8, head 11, of ViT-B/16 encoders at $224{\times}224$ resolution and average over query positions to measure the attention received by each patch, without a \texttt{[cls]} token. The $14{\times}14$ map is normalised to $[0,1]$ per image and rendered with the \texttt{inferno} colormap.

\paragraph{Uncertainty.}
We compute $U_p$ (\Cref{eq:uncertainty}) from 50 prior draws with the full image as context. Error maps mask each $5{\times}5$-patch block (stride 3), measure mean squared error between the mean prediction and the layer-normalised target over embedding dimensions, and average overlapping blocks. Maps are normalised to $[0,1]$, upsampled, and overlaid using \texttt{inferno} with $\alpha{=}0.75$. Quantitative evaluation uses pretraining \texttt{multi-block} masks and the same error definition: pooled and within-image Spearman correlations use 3000 ImageNet-1K validation images with $N{=}8$; the sample-count sweep uses 300 images with independently seeded draws. Calibration uses per-coordinate Gaussian predictions estimated from $N{=}50$ samples. We compute $\mathrm{ECE}_{\mathrm{reg}}$ as the mean absolute difference between nominal levels $0.05,0.10,\ldots,0.95$ and observed target coverage below the corresponding one-sided quantiles. A scalar temperature rescales predictive standard deviations, fitted by maximum likelihood on 500 images and evaluated on a disjoint 500-image split against a constant-variance baseline with the same mean.

\clearpage
\section{Ablation Study}
\label{sec:app-ablation}

We study \model on ImageNet-100 using ViT-B/16 trained for 400 epochs with batch size 512, varying one setting at a time. The default predictor has depth 6 and width 384, with $d_z{=}128$, $\beta{=}0.010$ with a linear warmup fraction of 0.3, and no free bits. We evaluate frozen average-pooled features using $k$-NN with $k{=}20$ and cosine distance~\citep{caron2021dino}, and $\ell_2$-regularised logistic regression on single center crops~\citep{assran2022msn}. Results average three independent pretraining seeds. Matched I-JEPA runs use the same optimiser, schedule, batch size, and training duration.

\subsection{Predictor Architecture}
\label{subsec:app-pred-arch}
We vary the latent dimension, predictor depth, and predictor width (\Cref{tab:latent_dim,tab:pred_depth,tab:pred_emb}). The latent dimension $d_z{=}128$ gives the best results on both metrics; increasing it to 256 or 512 does not improve performance. Shallower predictors improve $k$-NN accuracy, but depth 6 gives the strongest logistic regression result, showing that the preferred depth depends on the evaluation. Narrower predictors also improve $k$-NN, while width 192 achieves the highest logistic regression accuracy, slightly above width 384. We retain depth 6 and width 384 to match the I-JEPA baseline rather than selecting a separately tuned architecture for \model. The contrasting trends in the two metrics also caution against treating a single predictor setting as optimal for every downstream evaluation.

\subsection{KL Regularisation}
\label{subsec:app-kl-reg}
We vary the KL weight $\beta$, its warmup fraction, and the free-bits threshold $\lambda$ (\Cref{tab:kl_beta,tab:kl_warmup,tab:free_bits}). The default $\beta{=}0.010$ gives the strongest results; $\beta{=}0.050$ remains close, while larger weights reduce both metrics. A warmup fraction of 0.3 gives the best results; applying the final KL weight from the start performs worst, and warming up over the full schedule also underperforms. Free bits do not consistently improve both metrics: $\lambda{=}0.001$ increases $k$-NN accuracy but reduces logistic regression accuracy, and larger thresholds degrade both. We therefore use a warmup fraction of 0.3 and $\lambda{=}0$; background on free bits is given in Appendix~\ref{subsec:disc-freebits}.

\subsection{Masking Strategy}
\label{subsec:app-masking}
We compare five masking strategies (\Cref{table:ablation_masking}): \texttt{random} drops individual patches at a fixed ratio, as in MAE~\citep{he2021mae}; \texttt{grid} uses a fixed strided context with uniformly spaced targets; \texttt{rasterized} assigns three image quadrants as targets; \texttt{block} masks a single large rectangle following BERT/BEiT-style masking~\citep{devlin2018bert,bao2021beit}; and \texttt{multi-block} samples four target rectangles from a larger context block, as in I-JEPA~\citep{assran2023ijepa}. The \texttt{block} strategy performs best, followed by \texttt{multi-block}. We retain \texttt{multi-block} to match I-JEPA and hold the prediction task fixed. Relative gains depend on masking: \model improves both metrics under \texttt{multi-block}, \texttt{random}, and \texttt{rasterized}, while I-JEPA is stronger under \texttt{grid}; under \texttt{block} the methods have similar $k$-NN accuracy and \model improves logistic regression accuracy (\Cref{table:ablation_ijepa}). The results therefore support gains under shared masking settings rather than a uniform advantage across all strategies.

\subsection{Fair Comparison with I-JEPA \texorpdfstring{\citep{assran2023ijepa}}{}}
\label{subsec:app-ijepa-rerun}
To distinguish the effect of the stochastic predictor from architectural and training choices, we retrain I-JEPA across the same predictor-depth, predictor-width, and masking sweeps under the matched protocol above (\Cref{table:ablation_ijepa}). \model improves both metrics at most tested settings, including predictor depths 6--12 and widths 192--768. I-JEPA remains stronger with depths 2 and 4, width 96, and grid masking. Thus, the gains are not universal, and some differences are small relative to seed variation. We keep settings shared between methods and do not separately tune masking, so each comparison isolates the predictor under the same task.

\begin{table*}[t]
\centering
\caption{{\bf Ablation study of \model} on ImageNet-100, ViT-B/16, 400 epochs, batch size 512. We report $k$-NN top-1 and logistic regression top-1 accuracy (\%). We {\fboxsep 1pt\colorbox{gray!15}{highlight}} the default hyperparameter.}
\label{table:ablation}
\setlength{\tabcolsep}{6pt}
\renewcommand{\arraystretch}{1.1}
\small
%
\subfloat[
\textbf{Predictor depth}.
\label{tab:pred_depth}
]{
\begin{minipage}[t]{0.3\linewidth}
\centering
\begin{tabular}{lcc}
\toprule
\textbf{\texttt{pred\_depth}} & \textbf{knn} & \textbf{logistic} \\
\midrule
2  & 50.9 & 61.5 \\
4  & 41.6 & 61.3 \\
\rowcolor{gray!15} \textbf{6}  & 38.7 & 63.9 \\
8  & 28.3 & 56.4 \\
12 & 26.1 & 54.3 \\
\bottomrule
\end{tabular}
\end{minipage}
}
\hspace{0.3em}
\subfloat[
\textbf{Predictor width}.
\label{tab:pred_emb}
]{
\begin{minipage}[t]{0.3\linewidth}
\centering
\begin{tabular}{lcc}
\toprule
\texttt{pred\_width} & \textbf{knn} & \textbf{logistic} \\
\midrule
96  & 47.9 & 61.9 \\
192 & 47.8 & 64.1 \\
\rowcolor{gray!15} \textbf{384} & 38.7 & 63.9 \\
504 & 36.2 & 62.7 \\
768 & 30.8 & 59.9 \\
\bottomrule
\end{tabular}
\end{minipage}
}
\hspace{0.3em}
\subfloat[
\textbf{Latent dimension $d_z$}.
\label{tab:latent_dim}
]{
\begin{minipage}[t]{0.3\linewidth}
\centering
\begin{tabular}{lcc}
\toprule
\texttt{latent\_dim} & \textbf{knn} & \textbf{logistic} \\
\midrule
32  & 32.5 & 61.4 \\
64  & 29.9 & 57.8 \\
\rowcolor{gray!15} \textbf{128} & 38.7 & 63.9 \\
256 & 29.6 & 59.5 \\
512 & 35.6 & 62.3 \\
\bottomrule
\end{tabular}
\end{minipage}
}
\\[10pt]
%
\subfloat[
\textbf{KL weight $\beta$}.
\label{tab:kl_beta}
]{
\begin{minipage}[t]{0.3\linewidth}
\centering
\begin{tabular}{lcc}
\toprule
\texttt{kl\_beta\_final} & \textbf{knn} & \textbf{logistic} \\
\midrule
0.001 & 34.9 & 59.9 \\
\rowcolor{gray!15} \textbf{0.010} & 38.7 & 63.9 \\
0.050 & 38.0 & 62.9 \\
0.100 & 34.1 & 55.3 \\
1.000 & 30.2 & 44.6 \\
\bottomrule
\end{tabular}
\end{minipage}
}
\hspace{0.3em}
\subfloat[
\textbf{KL warmup fraction}.
\label{tab:kl_warmup}
]{
\begin{minipage}[t]{0.3\linewidth}
\centering
\begin{tabular}{lcc}
\toprule
\texttt{kl\_warmup\_frac} & \textbf{knn} & \textbf{logistic} \\
\midrule
0.00 & 26.6 & 54.9 \\
0.10 & 29.0 & 56.4 \\
0.20 & 29.9 & 57.0 \\
\rowcolor{gray!15} \textbf{0.30} & 38.7 & 63.9 \\
1.00 & 33.8 & 59.4 \\
\bottomrule
\end{tabular}
\end{minipage}
}
\hspace{0.3em}
\subfloat[
\textbf{Free-bits threshold $\lambda$}.
\label{tab:free_bits}
]{
\begin{minipage}[t]{0.3\linewidth}
\centering
\begin{tabular}{lcc}
\toprule
\texttt{kl\_free\_bits} & \textbf{knn} & \textbf{logistic} \\
\midrule
\rowcolor{gray!15} \textbf{0.000} & 38.7 & 63.9 \\
0.001 & 40.1 & 63.4 \\
0.010 & 35.5 & 61.7 \\
0.100 & 25.2 & 52.5 \\
0.500 & 30.0 & 57.7 \\
\bottomrule
\end{tabular}
\end{minipage}
}
\end{table*}
\begin{table*}[t]
\centering
\caption{{\bf Masking strategy ablation of \model} on ImageNet-100, ViT-B/16, 400 epochs, \model. We report $k$-NN top-1 and logistic regression top-1 accuracy (\%). We {\fboxsep1pt\colorbox{gray!15}{highlight}} the default strategy.}
\label{table:ablation_masking}
\small
\setlength{\tabcolsep}{3.7pt}
\begin{tabular}{llllllcc}
\toprule
& \multicolumn{3}{c}{\textbf{Targets}} & \multicolumn{2}{c}{\textbf{Context}} & & \\
\cmidrule(lr){2-4} \cmidrule(lr){5-6}
\textbf{Mask} & \textbf{Type} & \textbf{Freq.} & \textbf{Selection} & \textbf{Type} & \textbf{Avg.\ Ratio}$^*$ & \textbf{knn} & \textbf{logistic} \\
\midrule
\texttt{grid} & Strided & 1 & Fixed uniform & Grid(stride=2) & 0.25 & 26.9 & 38.2 \\
\texttt{random} & Random(0.6) & 1 & Random patches & Complement & 0.40 & 36.9 & 54.6 \\
\texttt{rasterized} & Quadrant & 3 & Fixed quadrant & Complement & 0.25 & 27.5 & 50.1 \\
\texttt{block} & Block(0.6) & 1 & Random rectangle & Complement & 0.40 & 58.4 & 67.9 \\
\rowcolor{gray!15}
\texttt{multi-block} & Block(0.15,0.2) & 4 & Random rectangle & Block(0.85,1.0)$\times$Complement & 0.25 & 38.7 & 63.9 \\
\bottomrule
\end{tabular}
\smallskip\\
\raggedright\footnotesize
$^*$Avg.\ Ratio is the average number of patches in the context block relative to the total number of patches in the image.
\end{table*}
\begin{table*}[t]
\centering
\caption{{\bf Ablation study of I-JEPA} on ImageNet-100, ViT-B/16, 400 epochs, batch size 512. We report $k$-NN top-1 and logistic regression top-1 accuracy (\%), along with $\Delta = \text{S-JEPA} - \text{I-JEPA}$ at the same hyperparameter value. We {\fboxsep 1pt\colorbox{gray!15}{highlight}} the default hyperparameter.}
\label{table:ablation_ijepa}
\setlength{\tabcolsep}{2pt}
\renewcommand{\arraystretch}{1.1}
\small
\subfloat[
\textbf{Predictor depth}.
\label{tab:ijepa_pred_depth}
]{
\begin{minipage}[t]{0.3\linewidth}
\centering
\begin{tabular}{lcccc}
\toprule
\texttt{pred\_depth} & \textbf{knn} & $\Delta$ & \textbf{logistic} & $\Delta$ \\
\midrule
2  & 53.9 & \textcolor{red!60!black}{-3.0} & 63.8 & \textcolor{red!60!black}{-2.3} \\
4  & 43.0 & \textcolor{red!60!black}{-1.4} & 62.5 & \textcolor{red!60!black}{-1.2} \\
\rowcolor{gray!15} 6  & 36.1 & \textcolor{green!60!black}{+2.6} & 61.1 & \textcolor{green!60!black}{+2.8} \\
8  & 26.6 & \textcolor{green!60!black}{+1.7} & 54.3 & \textcolor{green!60!black}{+2.1} \\
12 & 24.5 & \textcolor{green!60!black}{+1.6} & 49.7 & \textcolor{green!60!black}{+4.6} \\
\bottomrule
\end{tabular}
\end{minipage}
}
\hspace{0.3em}
\subfloat[
\textbf{Predictor width}.
\label{tab:ijepa_pred_emb}
]{
\begin{minipage}[t]{0.3\linewidth}
\centering
\begin{tabular}{lcccc}
\toprule
\texttt{pred\_width} & \textbf{knn} & $\Delta$ & \textbf{logistic} & $\Delta$ \\
\midrule
96  & 49.5 & \textcolor{red!60!black}{-1.6} & 62.3 & \textcolor{red!60!black}{-0.4} \\
192 & 47.2 & \textcolor{green!60!black}{+0.6} & 64.0 & \textcolor{green!60!black}{+0.1} \\
\rowcolor{gray!15} 384 & 36.1 & \textcolor{green!60!black}{+2.6} & 61.1 & \textcolor{green!60!black}{+2.8} \\
504 & 29.0 & \textcolor{green!60!black}{+7.2} & 56.8 & \textcolor{green!60!black}{+5.9} \\
768 & 26.6 & \textcolor{green!60!black}{+4.2} & 55.5 & \textcolor{green!60!black}{+4.4} \\
\bottomrule
\end{tabular}
\end{minipage}
}
\hspace{0.3em}
\subfloat[
\textbf{Masking strategy}.
\label{tab:ijepa_masking}
]{
\begin{minipage}[t]{0.3\linewidth}
\centering
\begin{tabular}{lcccc}
\toprule
\textbf{Mask} & \textbf{knn} & $\Delta$ & \textbf{logistic} & $\Delta$ \\
\midrule
\texttt{grid}        & 29.0 & \textcolor{red!60!black}{-2.1} & 45.9 & \textcolor{red!60!black}{-7.7} \\
\texttt{random}      & 36.6 & \textcolor{green!60!black}{+0.3} & 49.0 & \textcolor{green!60!black}{+5.6} \\
\texttt{rasterized}  & 22.8 & \textcolor{green!60!black}{+4.7} & 45.4 & \textcolor{green!60!black}{+4.7} \\
\texttt{block}       & 58.7 & \textcolor{red!60!black}{-0.3} & 66.1 & \textcolor{green!60!black}{+1.8} \\
\rowcolor{gray!15} \texttt{multi-block} & 36.1 & \textcolor{green!60!black}{+2.6} & 61.1 & \textcolor{green!60!black}{+2.8} \\
\bottomrule
\end{tabular}
\end{minipage}
}
\end{table*}

\subsection{Role of EMA and Stop-Gradient}
\label{subsec:app-ema}
We remove EMA by sharing weights between the context and target branches, then additionally remove stop-gradient (\Cref{tab:ema}). Alongside accuracy, effective rank on validation encoder features indicates how broadly the representation spans feature space. Without EMA, both methods lose accuracy and effective rank, with a larger drop for I-JEPA. \model retains more feature diversity in this setting, but removing stop-gradient as well reduces both methods to near-chance accuracy and very low effective rank. The stochastic predictor therefore does not remove the need for the target-encoder safeguards, and we retain both in the default configuration. Effective rank complements accuracy here: its decline shows that removing these safeguards changes the feature distribution as well as downstream performance. The remaining diversity without EMA should therefore be interpreted as greater robustness in this ablation, rather than evidence that EMA can be discarded without a cost.

\begin{table}[!t]
\centering
\caption{{\bf Role of EMA and stop-gradient (\texttt{sg}).} ImageNet-100, ViT-B/16, 400 epochs. Effective rank is a collapse diagnostic computed on $\ell_2$-normalised encoder features over the validation set; low values indicate representation collapse independently of accuracy. We {\fboxsep1pt\colorbox{gray!15}{highlight}} the default configuration.}
\label{tab:ema}
\setlength{\tabcolsep}{1.5mm}
\begin{tabular}{lccccc}
\toprule
\textbf{Method} & \textbf{EMA} & \textbf{\texttt{sg}} & \textbf{knn} & \textbf{logistic} & \textbf{eff. rank} \\
\midrule
\multirow{3}{*}{I-JEPA~\citep{assran2023ijepa}}
 & \xmark & \xmark & 1.1 & 1.3 & 8 \\
 & \xmark & \cmark & 12.2 & 21.3 & 35 \\
 & \cmark & \cmark & 36.1 & 61.1 & 187 \\
[1ex]
\multirow{3}{*}{S-JEPA (ours)}
 & \xmark & \xmark & 1.1 & 1.2 & 6 \\
 & \xmark & \cmark & 22.4 & 41.7 & 76 \\
 & \cellcolor{gray!15}\cmark & \cellcolor{gray!15}\cmark & \cellcolor{gray!15}38.7 & \cellcolor{gray!15}63.9 & \cellcolor{gray!15}201 \\
\bottomrule
\end{tabular}
\end{table}

\newpage
\section{Additional Experiments}
\label{sec:app-extra}

\subsection{Encoder Visualisations}
\label{subsec:app-encoder-viz}

\Cref{fig:pca_vitb_appendix,fig:att_vitb_appendix} compare \model with publicly released ViT-B/16 checkpoints on six images using the protocols in Appendix~\ref{subsec:app-viz-protocols}. PCA highlights object and background structure, while attention maps show the regions attended to by a fixed head. We compare against DINO~\citep{caron2021dino}, iBOT~\citep{zhou2021ibot}, MAE~\citep{he2021mae}, MSN~\citep{assran2022msn}, and I-JEPA~\citep{assran2023ijepa}, covering view-invariance, pixel-reconstruction, and predictive pretraining objectives. Each row uses the same input image across methods to facilitate direct comparison of their spatial feature structure.

\xdef\encvizsec{\thesection}\xdef\encvizfig{\arabic{figure}}

\newpage
\subsection{Uncertainty Analysis}
\label{subsec:app-uncertainty}

\begin{table}[!t]
\centering
\caption{{\bf Uncertainty quality} of ViT-B/16 under the pretraining masks. Spearman correlations between $U_p$ and error on 3000 validation images ($N{=}8$); calibration on 500 held-out images ($N{=}50$), with the temperature $\tau$ fit on a disjoint split. NLL in parentheses: constant-variance baseline with the same mean. I-JEPA has $U_p{=}0$.}
\label{tab:unc_calibration}
\setlength{\tabcolsep}{2mm}
\begin{tabular}{lcc}
\toprule
 & \textbf{\model} & \textbf{I-JEPA} \\
\midrule
Pooled Spearman $\rho$ & 0.327 & -- \\
Within-image Spearman $\rho$ (mean) & 0.161 & -- \\
Images with within-image $\rho > 0$ & 76\% & -- \\
\midrule
$\mathrm{ECE}_{\mathrm{reg}}$, raw $\downarrow$ & 0.180 & 0.237 \\
$\mathrm{ECE}_{\mathrm{reg}}$, after $\tau$ $\downarrow$ & 0.068 & -- \\
NLL after $\tau$ $\downarrow$ & 0.63 (0.98) & -- \\
\bottomrule
\end{tabular}
\end{table}

\begin{table}[!t]
\centering
\caption{{\bf Sample count.} Pooled Spearman $\rho$ between $U_p$ and error, and $\mathrm{ECE}_{\mathrm{reg}}$ (Appendix~\ref{subsec:app-viz-protocols}), of \model ViT-B/16 against the number of prior draws $N$, on 300 validation images under the pretraining masks; each row is reseeded. I-JEPA has $U_p{=}0$.}
\label{tab:n_sweep}
\setlength{\tabcolsep}{3mm}
\begin{tabular}{lcc}
\toprule
$\bm{N}$ & \textbf{Spearman} $\bm{\rho}$ & $\mathbf{ECE_{reg}}$ $\downarrow$ \\
\midrule
5  & 0.300 & 0.184 \\
10 & 0.272 & 0.182 \\
25 & 0.289 & 0.181 \\
50 & 0.279 & 0.181 \\
\midrule
I-JEPA & -- & 0.237 \\
\bottomrule
\end{tabular}
\end{table}

Under pretraining masks, $U_p$ correlates with prediction error across pooled patches (Spearman $\rho{=}0.327$), but the mean within-image correlation is weaker ($\rho{=}0.161$, positive in 76\% of images; \Cref{tab:unc_calibration}). Under the visualisation protocol, mean within-image correlation is only $0.08$ across 412 images. Thus, $U_p$ indicates prediction difficulty across predictions but is not a reliable localiser of error within an image; the quantitative conclusions rely on masked evaluation.

Raw predictions are underdispersed, with $\mathrm{ECE}_{\mathrm{reg}}{=}0.180$ compared with $0.237$ for deterministic I-JEPA. Scalar recalibration reduces this to $0.068$ and gives a lower NLL than the constant-variance baseline (0.63 versus 0.98), indicating useful variation beyond the overall uncertainty scale (\Cref{tab:unc_calibration}). Increasing the sample count from 5 to 50 produces little change in calibration and no consistent improvement in correlation (\Cref{tab:n_sweep}). All protocols are described in Appendix~\ref{subsec:app-viz-protocols}.

\clearpage

\section{Broader Related Work}
\label{sec:app-related}

\subsection{Joint-Embedding Architectures}
\label{subsec:app-related-joint}

Self-supervised learning of visual representations with joint-embedding architectures is an active line of research~\citep{wu2018unsupervised,he2019moco,chen2020simclr,grill2020byol,chen2020mocov2,chen2020simsiam,caron2021dino,bardes2021vicreg,zhou2021ibot,assran2022msn}. These approaches train a pair of encoders to produce similar embeddings for two or more views of the same image. To avoid degenerate solutions, popular methods use explicit regularisation~\citep{chen2020simclr,caron2021dino,bardes2021vicreg,assran2021semi} or architectural constraints such as stop-gradient~\citep{chen2020simsiam}, momentum encoders~\citep{grill2020byol}, or asymmetric prediction heads~\citep{grill2020byol,chen2020simsiam,baevski2022data2vec}. Regularisation-based approaches typically maximise the volume of space occupied by the representations, motivated by the InfoMax principle~\citep{linsker1988self,ma2022principles}, often coupled with independence or low-dimensionality constraints~\citep{zbontar2021barlow,bardes2021vicreg}. Despite their proliferation, such approaches have been criticised for not discriminating between semantically meaningful information and noise~\citep{assran2022hidden}, and are typically biased towards global image representations, limiting applicability to dense prediction tasks~\citep{chen2022intra,gidaris2020learning,bardes2022vicregl}.

\subsection{Masked Prediction}
\label{subsec:app-related-masked}

An orthogonal line of work learns representations by masking parts of the input and predicting the hidden content~\citep{vincent2010stacked}. Autoregressive and denoising models predict missing inputs at the pixel level~\citep{chen2020gpt2,he2021mae} or at the patch token level using a learned tokeniser~\citep{bao2021beit,wei2021masked}. While these methods demonstrate impressive scalability, they typically learn features at a lower level of semantic abstraction compared to joint-embedding approaches~\citep{assran2022msn}. More recently, approaches combining joint-embedding and reconstruction objectives have been proposed~\citep{el2021large,zhou2021ibot}, adding local patch-level reconstruction losses to invariance-based pretraining to improve dense prediction performance. I-JEPA~\citep{assran2023ijepa} departs from both paradigms by predicting in \emph{latent} space rather than pixel space, yielding semantically richer representations without hand-crafted augmentations. Our work extends I-JEPA by replacing the deterministic predictor with a principled stochastic latent-variable model, directly fulfilling the intent of LeCun's original JEPA blueprint~\citep{lecun2022path}.

\subsection{Contrastive Predictive Coding}
\label{subsec:app-related-cpc}

Contrastive predictive coding~\citep{oord2018cpc} is closely related to masked prediction, using a contrastive objective to encourage patch representations to be predictive of neighbouring patches~\citep{henaff2020cpc}. In contrast to these contrastive approaches, \model is non-contrastive and does not discriminate between image patches; the goal is to predict the representations of target blocks from a context block via a stochastic predictor conditioned on the latent variable $\bz$.

\subsection{Probabilistic and World Models}
\label{subsec:app-related-world}

The VAE~\citep{kingma2013vae} and CVAE~\citep{sohn2015cvae} provide the foundational framework for stochastic latent-variable inference used in \model. World-model research has emphasised learning latent dynamics for prediction and planning~\citep{ha2018worldmodels,bowman2016generating}, motivating the use of stochastic latent variables to model multi-modal future distributions. Our work brings this perspective into the masked image prediction setting, operating entirely in representation space without any reconstruction objective, preserving the non-generative character of JEPA~\citep{lecun2022path,assran2023ijepa}.
\section{Discussion}
\label{sec:app-discussion}

\noindent This appendix positions \model against the works closest to it on each axis of its design space: the treatment of $\bz$ (\Cref{subsec:disc-stop,subsec:disc-littwin}), the JEPA objective itself (\Cref{subsec:disc-lejepa}), concurrent variational formulations of JEPA (\Cref{subsec:disc-varjepa,subsec:disc-vjepa,subsec:disc-vje}), latent dynamics models (\Cref{subsec:disc-rssm}), probabilistic embeddings (\Cref{subsec:disc-probemb}), alternative sources of predictive uncertainty (\Cref{subsec:disc-uncertainty}), learned priors in variational inference (\Cref{subsec:disc-vamp}), and posterior collapse (\Cref{subsec:disc-freebits}). No prior or concurrent work combines a stochastic content-level $\bz$, a context-conditioned learned prior, and the absence of any decoder, which is the combination \model introduces.

\subsection{Comparison with Prior Work of Bar et al. \texorpdfstring{\cite{bar2024stochastic}}{}}
\label{subsec:disc-stop}

\noindent \citet{bar2024stochastic} observe that masked prediction requires the right content \emph{at the right location}, and that the location is itself uncertain: from a partial view of a dog one may infer that a tail is present without being able to place it. They therefore make the positional embedding of each mask token a random variable, drawn from a Gaussian centred at the deterministic embedding with a learned low-rank covariance.

\begin{definition}{Stochastic Positional Embeddings (StoP)}{stop}
For mask token $j$ with deterministic positional embedding $\boldsymbol{\psi}_j$,
\begin{equation}
    \hat{\boldsymbol{\psi}}_j \sim \mathcal{N}\!\left(\boldsymbol{\psi}_j,\, \Sigma\right), \qquad \Sigma = A A^\top,
    \tag{StoP}
\end{equation}
with $A$ learned and shared across all mask tokens and all images. Context tokens keep deterministic positions, the prediction loss is that of Def.~\ref{def:jepa}, and there is no posterior, no ELBO, and no conditional prior.
\end{definition}

\noindent This is why we scope our claim to the \emph{content} of $\bz$ rather than asserting that $\bz$ has been deterministic in every prior instantiation (Sections~\ref{sec:introduction} and~\ref{sec:related}). StoP randomises \emph{where} $g_\phi$ is queried; \model models \emph{what} the occluded region contains. Three structural differences follow. First, the source of the distribution: $\Sigma$ is a single global parameter, whereas $\bmu_\theta(\sx)$ and $\bsig_\theta(\sx)$ are functions of the image and can therefore express that one image is ambiguous where another is determined. Second, the training signal: StoP has no target-conditioned inference network, so nothing ties the injected noise to the actual content of the masked region, whereas $\post$ is trained to explain $\sy$ and the KL term is what prevents it short-cutting prediction (\Cref{lemma:ib}). Third, the behaviour at inference: perturbing positions changes the query, whereas drawing $\bz \sim \prior$ produces distinct semantic completions of the same query, which is what makes the uncertainty maps and completions of Figures~\ref{fig:uncertainty} and~\ref{fig:rcdm} possible at all. The two mechanisms act on different arguments of $g_\phi(\sx, \bz)$ and are complementary rather than competing; a predictor stochastic in both position and content is a natural extension of either.

\subsection{Comparison with Prior Work of Littwin et al. \texorpdfstring{\cite{littwin2024enhancingjepas}}{}}
\label{subsec:disc-littwin}

\noindent \citet{littwin2024enhancingjepas} address the same difficulty from the opposite direction. Because information in natural images has a strong spatial bias, the feasibility of predicting a target window from a context window depends on their relative positions, and encoders that see only their own window cannot adapt to this. They therefore condition each encoder on the positions of the opposing window.

\begin{definition}{Spatial Conditioning}{spatialcond}
With $P_x$ and $P_y$ the positions of the context and target windows,
\begin{equation}
    \sx = f_x(\bx, P_y), \qquad \sy = f_y(\by, P_x),
    \tag{Spatial Conditioning}
\end{equation}
so that each encoder can modulate the features it emits according to the window it is asked to predict from or to. The predictor, the auxiliary variable, and the loss are those of Def.~\ref{def:jepa}.
\end{definition}

\noindent The relationship is one of division of labour rather than overlap. Spatial conditioning acts on the encoders and changes what $\sx$ and $\sy$ contain; \model leaves the encoders untouched and changes what conditions the predictor beyond $\sx$. More importantly, the two remove different residues of uncertainty. Conditioning on window positions removes the uncertainty that arises because the encoder does not know the geometry of the prediction problem, which is information the model could in principle have been given. It cannot remove the uncertainty that remains once the geometry is known, since the content of an occluded region is under-determined by the visible context however well the layout is understood. That irreducible component is what $\bz$ is introduced to carry, and conditional encoders and a stochastic latent are therefore composable.

\subsection{Comparison with LeJEPA of Balestriero and LeCun \texorpdfstring{\cite{balestriero2025lejepa}}{}}
\label{subsec:disc-lejepa}

\noindent In LeCun's JEPA blueprint the predictor maps $(\sx, \bz) \mapsto \hsy$, where $\bz$ captures the residual uncertainty about the target that is not deducible from the source. This construction presupposes an information asymmetry: in masked or temporal prediction the source is a partial observation, many target embeddings are consistent with a given $\sx$, and $\bz$ is what parameterises the resulting multimodality. \citet{balestriero2025lejepa} eliminate this asymmetry, replacing prediction with a pure invariance objective over symmetric augmentations and handling collapse through an explicit distributional regulariser rather than architectural heuristics.

\begin{definition}{LeJEPA}{lejepa}
Given $V$ views of sample $n$ with embeddings $\mathbf{z}_{n,v}$ and centroid $\boldsymbol{\mu}_n = \frac{1}{V_g}\sum_{v=1}^{V_g} \mathbf{z}_{n,v}$,
\begin{equation}
    \mathcal{L}_{\mathrm{pred}} = \frac{1}{V}\sum_{v'=1}^{V} \left\| \boldsymbol{\mu}_n - \mathbf{z}_{n,v'} \right\|_2^2,
    \tag{LeJEPA}
\end{equation}
regularised by SIGReg, which drives the embedding distribution toward $\mathcal{N}(\mathbf{0}, \mathbf{I})$ by characteristic-function matching along random one-dimensional projections. There is no predictor and no auxiliary variable.
\end{definition}

\noindent \model is not a competing instantiation of the same objective but a model of structure that LeJEPA removes by construction. Because all views are symmetric augmentations of the same image and no information gap separates them, the mapping from any view to the target centroid is deterministic; were an auxiliary variable to be injected into such a mapping, the optimal solution under \Cref{def:lejepa} would be to ignore it entirely. A structured latent with a KL regulariser therefore has no role to play: the anti-collapse function it would serve in a conditional VAE framing is discharged instead by SIGReg. LeJEPA trades the generative, multimodal structure of the full JEPA blueprint for a simpler invariance-plus-distributional-matching objective, which makes $\bz$ both unnecessary and degenerate under its loss landscape. The two methods are therefore best read as occupying different points in the design space rather than as strict alternatives, and the empirical comparison we report in Section~\ref{sec:experiments} should be read as a comparison of the representations each objective produces at a given architecture, not as evidence that one objective subsumes the other. Note that LeJEPA is a multi-view objective in the sense of \Cref{def:lejepa}: each image contributes $V$ views, in the standard configuration two global crops at $224^2$ and six local crops at $98^2$, and every one of them passes through a gradient-bearing encoder. A single LeJEPA epoch therefore pushes 608 encoder tokens per image against roughly 245 for S-JEPA, a factor of 2.5, and 12$\times$ as many gradient-bearing tokens, since our target encoder is EMA-updated and only the context block carries gradients. Over the full schedules of \Cref{table:linear} (600 vs.\ 100 epochs), S-JEPA processes ${\approx}2.4\times$ as many encoder tokens as LeJEPA, while LeJEPA processes ${\approx}2\times$ as many gradient-bearing tokens.

\subsection{Comparison with Concurrent Work of G\"ogl and Yau \texorpdfstring{\cite{gogl2026varjepa}}{}}
\label{subsec:disc-varjepa}

\noindent \citet{gogl2026varjepa} argue that the separation between JEPA and probabilistic generative modelling is largely rhetorical rather than structural, since the coupled encoder--predictor design mirrors the variational posteriors and learned conditional priors that arise from variational inference applied to a class of coupled latent-variable models. They reverse-engineer a probabilistic model whose ELBO yields the JEPA predictor, make that latent structure explicit through a single bound, and instantiate it for tabular data.

\begin{definition}{Var-JEPA}{varjepa}
A single ELBO over a coupled latent-variable model in which the encoder embeddings are the generative objects and the predictor is interpreted as a decoder,
\begin{equation}
    \mathcal{L}_{\mathrm{Var}} = \mathbb{E}_{q}\!\left[\log p(\mathrm{obs} \mid \cdot)\right] - \KL\!\left(q \;\|\; p\right),
    \tag{Var-JEPA}
\end{equation}
with observation-space decoders and reconstruction terms retained deliberately, on the grounds that reconstruction regularises the latents toward encoding predictive information.
\end{definition}

\noindent \model shares the variational perspective but differs in three respects. First, the setting: Var-JEPA is instantiated on tabular data and treats the embeddings themselves as the generative objects, while \model addresses spatial masked image prediction, where multimodality is not a modelling convenience but a property of the task, since many distinct completions of an occluded region are semantically valid. Second, the role of the latent variable: in Var-JEPA the latent code is obtained by reverse-engineering a generative model whose ELBO recovers an existing predictor, so $\bz$ functions as an interpretive device for a fixed architecture; in \model, $\bz$ is an explicit stochastic variable injected into the predictor at training time, with a learned context-conditioned prior used to draw samples at inference. Third, and most importantly, the direction of travel: Var-JEPA introduces observation-space generative structure in order to bridge JEPA and generative modelling, which is its stated goal, whereas \model deliberately moves the other way, keeping the CVAE entirely at the level of representations with no decoders and no reconstruction, so that the non-generative character of JEPA is preserved by design.

\subsection{Comparison with Concurrent Work of Huang \texorpdfstring{\cite{huang2026vjepa}}{}}
\label{subsec:disc-vjepa}

\noindent \citet{huang2026vjepa} introduce a probabilistic generalisation that learns a predictive distribution over \emph{future} latent states via a variational objective. The motivation is stochastic sequential control: the resulting model is connected to predictive state representations and Bayesian filtering, and its representations are shown to serve as sufficient information states for optimal control without pixel reconstruction.

\begin{definition}{VJEPA}{vjepa}
For latent states $s_t$ and a horizon step $t \to t{+}1$,
\begin{equation}
    \mathbb{E}_{q(\bz \mid s_t, s_{t+1})}\!\left[\log p(s_{t+1} \mid s_t, \bz)\right] - \KL\!\left(q(\bz \mid s_t, s_{t+1}) \;\|\; \mathcal{N}(\mathbf{0}, \mathbf{I})\right),
    \tag{VJEPA}
\end{equation}
where the prior over $\bz$ is a \emph{fixed} isotropic Gaussian rather than a learned function of the conditioning state.
\end{definition}

\noindent \model differs on all three axes that define its design space. First, the task: VJEPA addresses temporal, sequential prediction in a control setting, while \model addresses spatial masked image prediction. Second, the prior: VJEPA uses a fixed $\mathcal{N}(\mathbf{0}, \mathbf{I})$, whereas \model employs a context-conditioned learned prior $\prior$ trained to anticipate the distribution of plausible completions from visible patches alone. This distinction is not cosmetic. A fixed prior samples from an uninformed distribution at inference and therefore cannot express that one image is ambiguous where another is determined; it also changes the geometry of the objective, since the KL can then be reduced only by drawing each posterior toward a common uninformative distribution, whereas a learned prior can instead move toward the posteriors it must match (\Cref{subsec:remark-prior}). Third, the role of $\bz$: in VJEPA it captures temporal uncertainty over future states for planning, whereas in \model it captures spatial semantic ambiguity over an occluded region, which is what gives rise to the patch-level uncertainty maps and the diverse completions of Figures~\ref{fig:uncertainty} and~\ref{fig:rcdm}.

\subsection{Comparison with Concurrent Work of Oji and Fieguth \texorpdfstring{\cite{oji2026joint}}{}}
\label{subsec:disc-vje}

\noindent \citet{oji2026joint} introduce a reconstruction-free non-contrastive framework that places a latent-variable model directly on paired encoder embeddings. They maximise a symmetric conditional ELBO whose likelihood is a heavy-tailed Student-$t$ distribution on a polar decomposition of the target embedding, separating directional agreement from magnitude consistency, with an amortised inference network whose feature-wise posterior variances are shared with the likelihood scale to capture anisotropic uncertainty.

\begin{definition}{VJE}{vje}
For embeddings $(\mathbf{u}, \mathbf{v})$ produced by the \emph{same} encoder from two augmented views,
\begin{equation}
    \tfrac{1}{2}\!\!\sum_{(a,b),(b,a)}\!\! \left( \mathbb{E}_{q(\bz \mid a, b)}\!\left[\log p_{\mathrm{St}}(b \mid a, \bz)\right] - \KL\!\left(q(\bz \mid a, b) \;\|\; p(\bz)\right) \right),
    \tag{VJE}
\end{equation}
with $p_{\mathrm{St}}$ a Student-$t$ likelihood on the polar decomposition of $b$. There is no predictor, no masking, and no context--target asymmetry.
\end{definition}

\noindent The surface similarity, operation in representation space under a variational ELBO with no pixel reconstruction, conceals a fundamental architectural difference. VJE is symmetric: both views pass through the same encoder and the bound is maximised symmetrically over the pair, so the latent variable models the shared embedding uncertainty of two augmentations of the same image. \model is asymmetric: $\bz$ is conditioned on the context representation through the prior and parameterises the distribution over semantically valid completions of a region that was never observed, which is a different quantity from view-level embedding uncertainty. The two methods therefore address different problems with different inductive structures, VJE being positioned as a probabilistic alternative to BYOL and VICReg-style invariance learning while \model is positioned as a stochastic extension of I-JEPA-style masked prediction.

\subsection{Relation to Latent Dynamics Models of Hafner et al. \texorpdfstring{\cite{hafner2019learning,hafner2020dreamcontrol}}{}}
\label{subsec:disc-rssm}

\noindent The closest structural precedent for the prior--posterior asymmetry we adopt lies outside self-supervised vision, in latent-dynamics world models. \citet{hafner2019learning} split the latent state into deterministic and stochastic components precisely so that the model can represent multiple futures rather than a single rollout, and train a transition prior that predicts the next state from the previous one alongside a filtering posterior that additionally observes the incoming frame. \citet{hafner2020dreamcontrol} extend this to behaviour learning by imagining trajectories in the learned latent space.

\begin{definition}{Recurrent State-Space Model (RSSM)}{rssm}
With deterministic path $h_t$, stochastic state $s_t$, and observation $o_t$, the model couples
\begin{equation}
    \underbrace{p(s_t \mid h_t)}_{\text{transition prior}}, \qquad \underbrace{q(s_t \mid h_t, o_t)}_{\text{filtering posterior}}, \qquad \underbrace{p(o_t \mid h_t, s_t)}_{\text{observation model}},
    \tag{RSSM}
\end{equation}
trained by maximising a variational bound in which the observation model reconstructs the raw frame $o_t$.
\end{definition}

\noindent We adopt the same asymmetry, with $\prior$ playing the role of the transition prior and $\post$ that of the filtering posterior, and for the same reason: the prior must learn to anticipate what the posterior sees, which is what makes prior samples meaningful at inference. The difference is the third component. The RSSM closes the loop through an observation model that reconstructs pixels, which is what supplies its learning signal and what makes it a generative model; \model has no such term, and the prediction signal arrives entirely from the frozen target representation $\sy$. Our bound is consequently on $\log p(\sy \mid \sx)$ rather than on an observation likelihood, and the quantity modelled is semantic content in a learned embedding space rather than sensory reconstruction. The setting differs as well: latent-dynamics models resolve uncertainty over time in a controlled environment, whereas the ambiguity \model addresses is spatial and static, present in a single image with no subsequent observation to disambiguate it.

\subsection{Relation to Probabilistic Embeddings of Oh et al. \texorpdfstring{\cite{oh2019modeling}}{} and Chun et al. \texorpdfstring{\cite{chun2021probabilistic}}{}}
\label{subsec:disc-probemb}

\noindent Representing an embedding as a distribution rather than a point has an established history in metric learning and retrieval. \citet{oh2019modeling} map each input to a random variable trained under the variational information bottleneck and obtain a per-example uncertainty. \citet{chun2021probabilistic} make the argument in the cross-modal setting, observing that deterministic functions cannot represent one-to-many correspondences between images and captions, and that a distributional embedding both improves retrieval and yields interpretable uncertainty.

\begin{definition}{Probabilistic Embedding (HIB, PCME)}{probemb}
An input $\bx$ is mapped not to a point but to a distribution over the embedding space,
\begin{equation}
    p(\mathbf{e} \mid \bx) = \mathcal{N}\!\left(\bmu(\bx),\, \mathrm{diag}(\bsig^2(\bx))\right),
    \tag{PE}
\end{equation}
with matching performed in expectation over samples and $\bsig(\bx)$ interpreted as the ambiguity of the \emph{observed} input $\bx$.
\end{definition}

\noindent We share with this line the position that a point estimate is inadequate when the mapping is genuinely one-to-many. The distinction is what the distribution is over. In probabilistic embeddings the uncertainty attaches to an input the model has seen, and $\bsig(\bx)$ expresses that $\bx$ itself is blurred, occluded, or polysemous. In \model the input is not ambiguous at all: the context patches are clean and fully observed, and the uncertainty attaches to a region that was deliberately withheld. The distribution is therefore predictive rather than perceptual, defined over completions of something absent rather than interpretations of something present, and it is conditioned on the visible context through a prior that must anticipate targets it never sees. This is also why the uncertainty is spatially resolved in our setting, varying across positions within a single image, whereas a probabilistic embedding assigns a single uncertainty to the input as a whole.

\subsection{Alternative Sources of Predictive Uncertainty}
\label{subsec:disc-uncertainty}

\noindent A deterministic I-JEPA predictor yields $U_p = 0$ everywhere, but it does not follow that no deterministic architecture can produce a per-patch uncertainty signal, and we state the comparison precisely rather than categorically. Three standard mechanisms apply to a JEPA predictor with no latent variable: Monte Carlo dropout \citep{gal2016dropout}, which interprets test-time dropout as approximate Bayesian inference; deep ensembles \citep{lakshminarayanan2017simple}, which take the variance across independently trained predictors; and a heteroscedastic output head \citep{kendall2017uncertainties}, which regresses a per-patch variance directly alongside the mean.

\begin{definition}{Deterministic Uncertainty Baselines}{uncbaselines}
For a predictor $g$ and patch $p$, a per-patch variance can be obtained as
\begin{equation}
    U_p^{\mathrm{MC}} = \mathrm{Var}_{i}\!\left[g_{\phi, \epsilon_i}^{(p)}\right], \quad
    U_p^{\mathrm{ens}} = \mathrm{Var}_{m}\!\left[g_{\phi_m}^{(p)}\right], \quad
    U_p^{\mathrm{het}} = \big(\bsig_\phi(\sx)^{(p)}\big)^2,
    \tag{Deterministic Uncertainty}
\end{equation}
with $\epsilon_i$ independent dropout masks, $\phi_m$ independently trained predictors, and $\bsig_\phi$ a learned variance head. None requires a latent variable.
\end{definition}

\noindent The distinction we draw is therefore not that \model produces a variance where these cannot, but what the variance is \emph{about} and what else it yields. Dropout and ensembles measure epistemic uncertainty, reducible with more data and not tracking the intrinsic ambiguity of the masked region, and ensembles require $M$ independent pretraining runs. A heteroscedastic head is the strongest of the three, capturing aleatoric uncertainty per patch, but emits a scalar variance with no distribution over completions. In \model the uncertainty arises from the same latent that generates the completions of Figure~\ref{fig:rcdm}, which is why the two capabilities follow from a single mechanism rather than from two separate additions to the architecture.

\subsection{Relation to Learned Priors of Tomczak and Welling \texorpdfstring{\cite{tomczak2018vaevampprior}}{}}
\label{subsec:disc-vamp}

\noindent The result established in \Cref{subsec:remark-prior}, that the optimal prior for a fixed posterior is the aggregate posterior, is the conditional analogue of a known result rather than a new one. \citet{tomczak2018vaevampprior} identify the aggregated posterior as the prior that maximises the VAE ELBO, connect it to the information bottleneck, and observe that using it directly overfits, which motivates approximating it by a mixture of posteriors evaluated at a small number of learnable pseudo-inputs.

\begin{definition}{Optimal Prior and the VampPrior}{vamp}
For a fixed posterior $q_\phi$, the ELBO-maximising prior over the dataset is the aggregated posterior
\begin{equation}
    p^*(\bz) = \frac{1}{N}\sum_{i=1}^{N} q_\phi(\bz \mid \bx_i),
    \qquad\text{approximated by}\qquad
    p(\bz) = \frac{1}{K}\sum_{k=1}^{K} q_\phi(\bz \mid u_k),
    \tag{Vamp}
\end{equation}
with $u_1, \ldots, u_K$ learnable pseudo-inputs and $K \ll N$.
\end{definition}

\noindent We claim neither the result nor the observation that it bears on posterior collapse. What is specific to \model is the conditional form and its structural consequence. Our aggregation is not over the dataset but over the targets consistent with a given context, $q_\psi^{\mathrm{agg}}(\bz \mid \sx) = \mathbb{E}_{\sy \mid \sx}[\post]$, so the optimal prior is a different distribution for every image and encodes the set of plausible completions of that image specifically. This is also why the pseudo-input approximation is unnecessary here: the prior is amortised as a function of $\sx$ by a small network rather than represented as an explicit mixture, so the overfitting that motivates the VampPrior does not arise in the same form. The mechanism this makes available, reducing the divergence by moving the prior toward the posteriors rather than by drawing every posterior onto a common uninformative distribution, is precisely what a fixed prior forecloses. The optimum of \Cref{subsec:remark-prior} is a mixture, which a diagonal Gaussian prior approximates rather than represents, so the multimodality in Figure~\ref{fig:rcdm} arises from the predictor's nonlinear map of a unimodal latent; a mixture prior or a flow posterior would represent it directly.

\subsection{Posterior Collapse and Free Bits of Kingma et al. \texorpdfstring{\cite{kingma2016iaf}}{}}
\label{subsec:disc-freebits}

\noindent Posterior collapse \citep{bowman2016generating,lucas2019dont} is a well-known failure mode of VAE-style objectives, in which the posterior converges to the prior and the latent variable carries no information beyond what the prior already encodes. \citet{kingma2016iaf} address it with free bits, which reserve a minimum KL budget for every latent dimension by detaching gradients for any dimension whose divergence falls below a threshold, guaranteeing that no single dimension is permitted to switch off.

\begin{definition}{Posterior Collapse and Free Bits}{collapse}
The latent $\bz$ collapses when $\post = \prior$ for almost all $(\sx, \sy)$, in which case $I_{q_\psi}(\bz; \sy \mid \sx) = 0$ by \Cref{lemma:ib} and the predictor is functionally deterministic. Free-bits regularisation enforces a per-dimension lower bound $\lambda > 0$:
\begin{equation}
    \mathcal{L}_{\mathrm{KL}}^{\lambda} = \sum_{k=1}^{d_z} \max\!\left( \KL^{(k)}\!\left(\post \;\|\; \prior\right),\; \lambda \right).
    \tag{FB}
    \label{eq:freebits}
\end{equation}
\end{definition}

\noindent We do not adopt this mechanism, for structural rather than empirical reasons. In a standard VAE the decoder is the only path from $\bz$ to the output, so forcing dimensions active necessarily improves reconstruction and free bits aligns with the objective. In \model the predictor also sees $\sx$, a strong alternative route to the target, so forcing $\bz$ to carry more than the predictor finds useful injects noise rather than structure. What keeps $\bz$ informative is instead the context-conditioned prior of \Cref{subsec:disc-vamp}: since the optimal prior is the aggregate posterior, the divergence can be reduced by moving $\prior$ toward the posteriors rather than collapsing them, and no per-dimension floor is required. We therefore adopt the plain objective of Def.~\ref{def:sjepa} as the default. Note that \Cref{def:collapse} concerns collapse of the \emph{latent}, distinct from that of the \emph{representation} discussed in Section~\ref{sec:related}: the KL bounds the information $\bz$ carries about $\sy$ and prevents it acting as a shortcut, but constrains nothing about the encoder outputs, which is why the EMA target encoder is retained as in I-JEPA \citep{assran2023ijepa}.

\FloatBarrier
\renewcommand{\thefigure}{\encvizsec.\arabic{figure}}
\setcounter{figure}{\encvizfig}
\begin{figure*}[ht]
\centering
\setlength{\tabcolsep}{1.0pt}
\renewcommand{\arraystretch}{1}
\small
\begin{tabular}{ccccccc}
Image & DINO & iBOT & MAE & MSN & I-JEPA & \model \\
[2pt]
\includegraphics[width=0.13\linewidth]{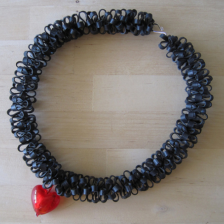} &
\includegraphics[width=0.13\linewidth]{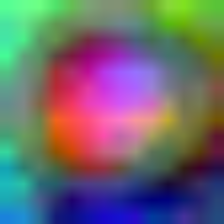} &
\includegraphics[width=0.13\linewidth]{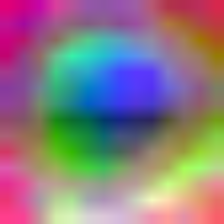} &
\includegraphics[width=0.13\linewidth]{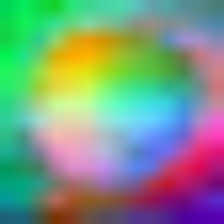} &
\includegraphics[width=0.13\linewidth]{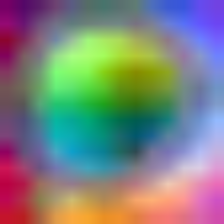} &
\includegraphics[width=0.13\linewidth]{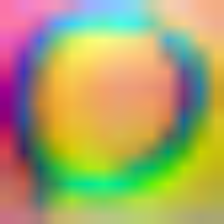} &
\includegraphics[width=0.13\linewidth]{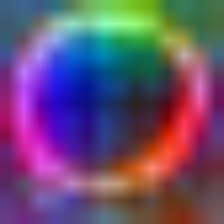} \\
[2pt]
\includegraphics[width=0.13\linewidth]{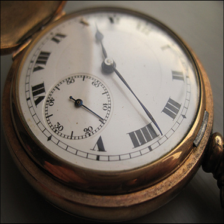} &
\includegraphics[width=0.13\linewidth]{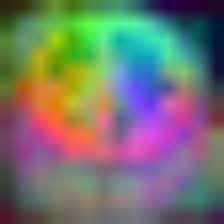} &
\includegraphics[width=0.13\linewidth]{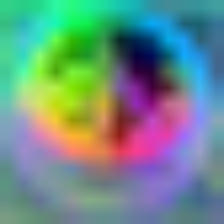} &
\includegraphics[width=0.13\linewidth]{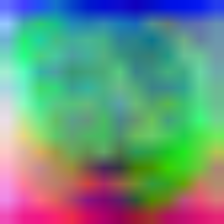} &
\includegraphics[width=0.13\linewidth]{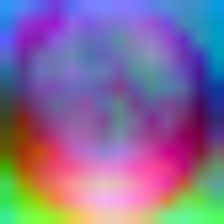} &
\includegraphics[width=0.13\linewidth]{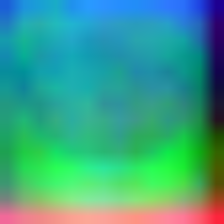} &
\includegraphics[width=0.13\linewidth]{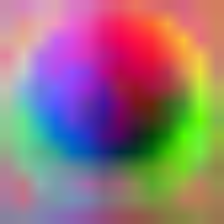} \\
[2pt]
\includegraphics[width=0.13\linewidth]{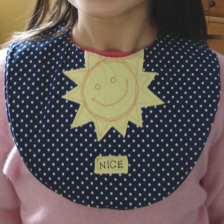} &
\includegraphics[width=0.13\linewidth]{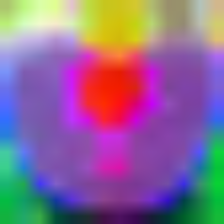} &
\includegraphics[width=0.13\linewidth]{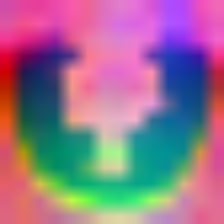} &
\includegraphics[width=0.13\linewidth]{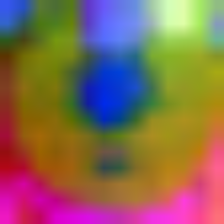} &
\includegraphics[width=0.13\linewidth]{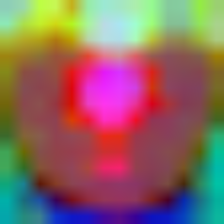} &
\includegraphics[width=0.13\linewidth]{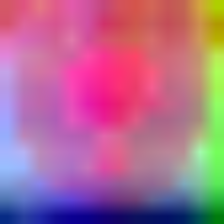} &
\includegraphics[width=0.13\linewidth]{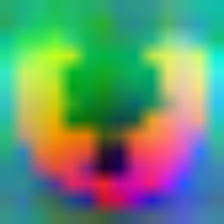} \\
[2pt]
\includegraphics[width=0.13\linewidth]{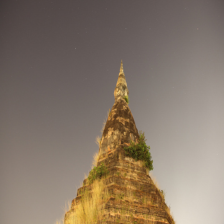} &
\includegraphics[width=0.13\linewidth]{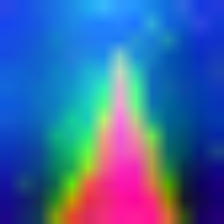} &
\includegraphics[width=0.13\linewidth]{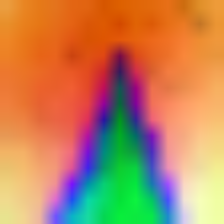} &
\includegraphics[width=0.13\linewidth]{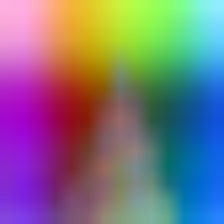} &
\includegraphics[width=0.13\linewidth]{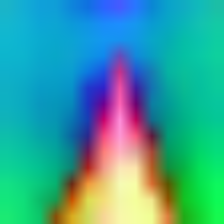} &
\includegraphics[width=0.13\linewidth]{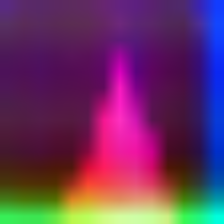} &
\includegraphics[width=0.13\linewidth]{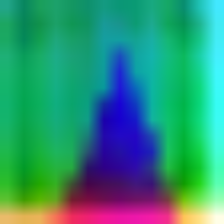} \\
[2pt]
\includegraphics[width=0.13\linewidth]{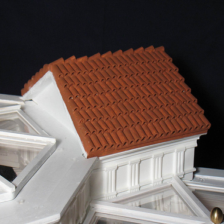} &
\includegraphics[width=0.13\linewidth]{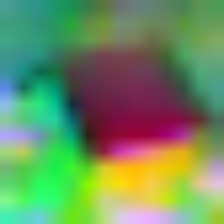} &
\includegraphics[width=0.13\linewidth]{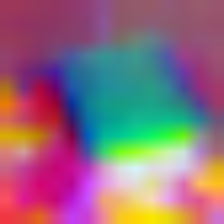} &
\includegraphics[width=0.13\linewidth]{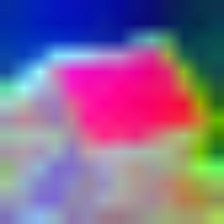} &
\includegraphics[width=0.13\linewidth]{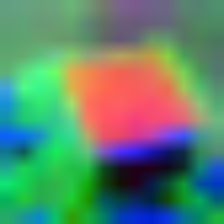} &
\includegraphics[width=0.13\linewidth]{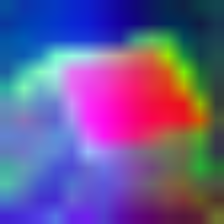} &
\includegraphics[width=0.13\linewidth]{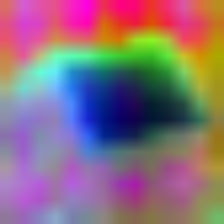} \\
[2pt]
\includegraphics[width=0.13\linewidth]{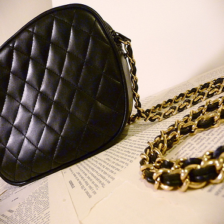} &
\includegraphics[width=0.13\linewidth]{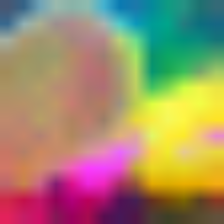} &
\includegraphics[width=0.13\linewidth]{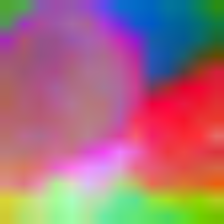} &
\includegraphics[width=0.13\linewidth]{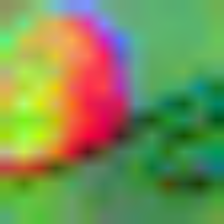} &
\includegraphics[width=0.13\linewidth]{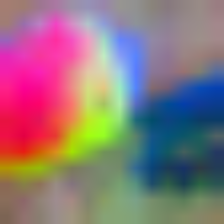} &
\includegraphics[width=0.13\linewidth]{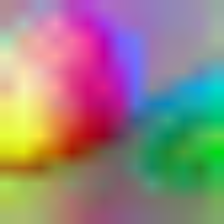} &
\includegraphics[width=0.13\linewidth]{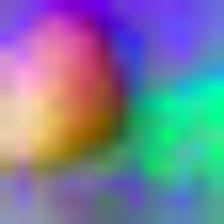} \\
\end{tabular}
\caption{{\bf PCA visualisations of encoder features (ViT-B/16).} Each row shows the original image followed by the PCA projection of patch features for DINO~\cite{caron2021dino} (400 epochs), iBOT~\cite{zhou2021ibot} (400 epochs), MAE~\cite{he2021mae} (1600 epochs), MSN~\cite{assran2022msn} (600 epochs), I-JEPA~\cite{assran2023ijepa} (600 epochs) and \model (600 epochs), all ViT-B/16 encoders pretrained on ImageNet-1K. All baseline visualisations are reproduced from publicly released checkpoints.}
\label{fig:pca_vitb_appendix}
\end{figure*}

\begin{figure*}[ht]
\centering
\setlength{\tabcolsep}{1.0pt}
\renewcommand{\arraystretch}{1}
\small
\begin{tabular}{ccccccc}
Image & DINO & iBOT & MAE & MSN & I-JEPA & \model \\
[2pt]
\includegraphics[width=0.13\linewidth]{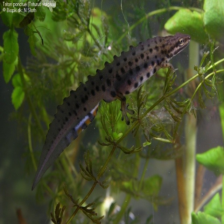} &
\includegraphics[width=0.13\linewidth]{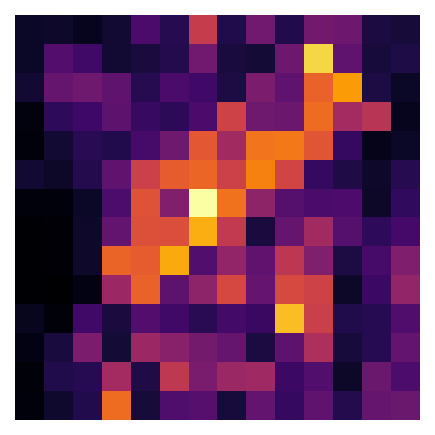} &
\includegraphics[width=0.13\linewidth]{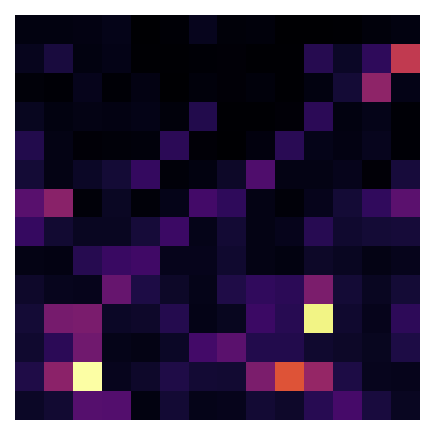} &
\includegraphics[width=0.13\linewidth]{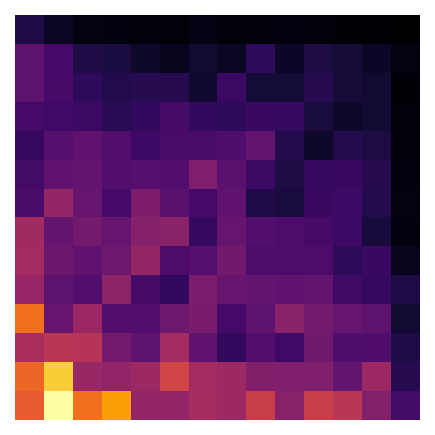} &
\includegraphics[width=0.13\linewidth]{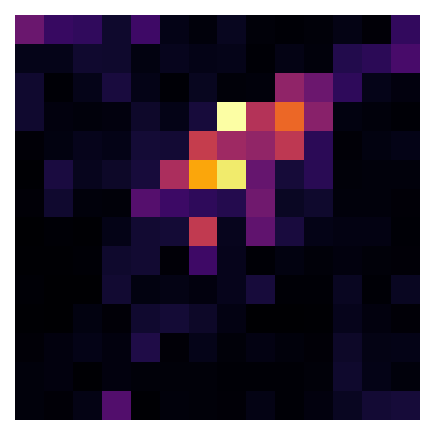} &
\includegraphics[width=0.13\linewidth]{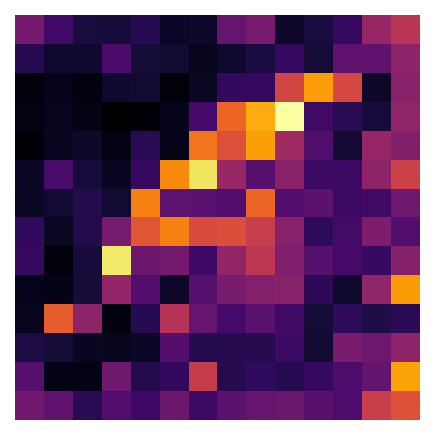} &
\includegraphics[width=0.13\linewidth]{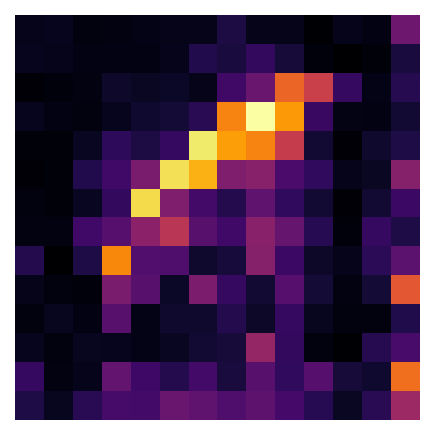} \\
[2pt]
\includegraphics[width=0.13\linewidth]{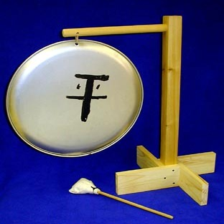} &
\includegraphics[width=0.13\linewidth]{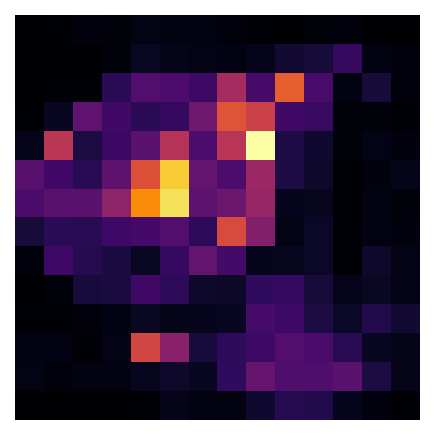} &
\includegraphics[width=0.13\linewidth]{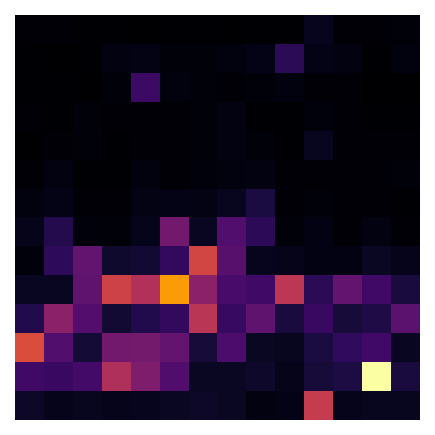} &
\includegraphics[width=0.13\linewidth]{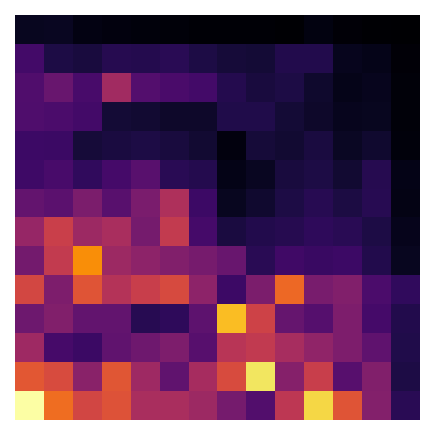} &
\includegraphics[width=0.13\linewidth]{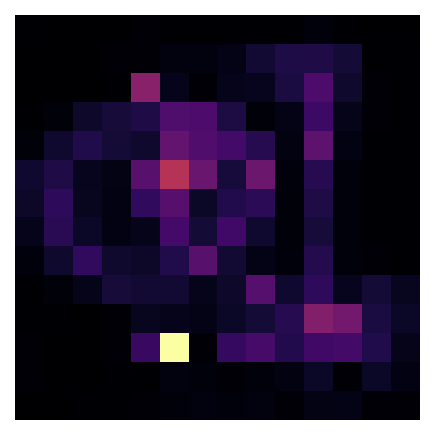} &
\includegraphics[width=0.13\linewidth]{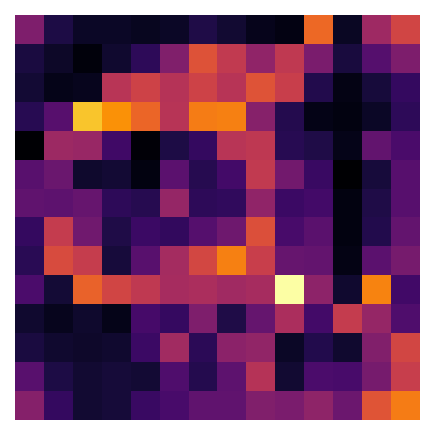} &
\includegraphics[width=0.13\linewidth]{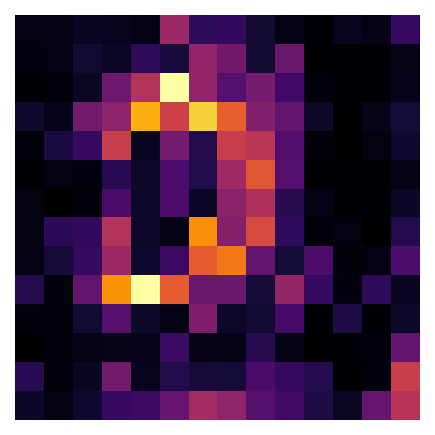} \\
[2pt]
\includegraphics[width=0.13\linewidth]{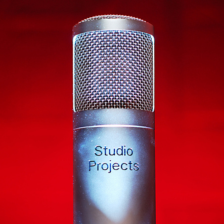} &
\includegraphics[width=0.13\linewidth]{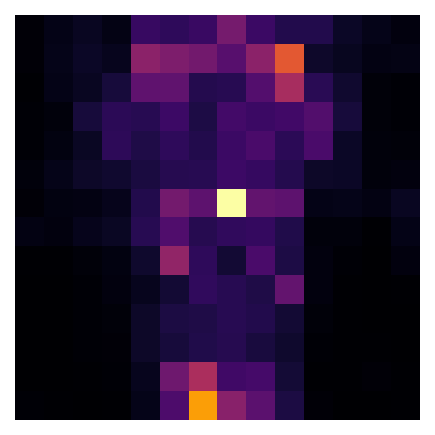} &
\includegraphics[width=0.13\linewidth]{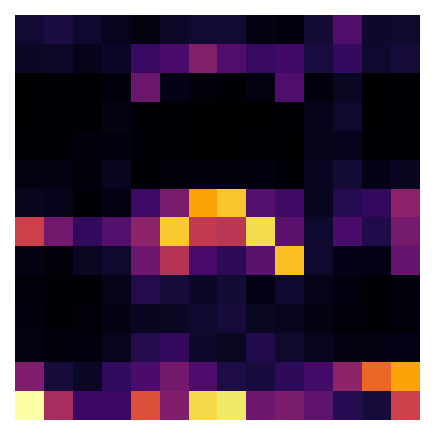} &
\includegraphics[width=0.13\linewidth]{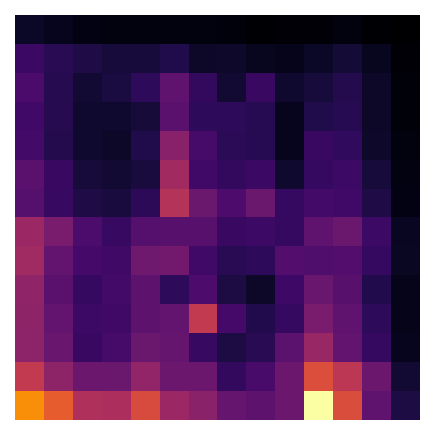} &
\includegraphics[width=0.13\linewidth]{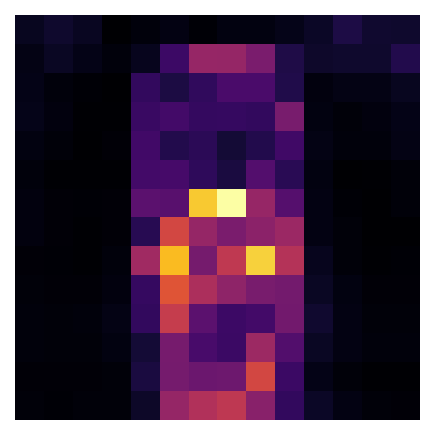} &
\includegraphics[width=0.13\linewidth]{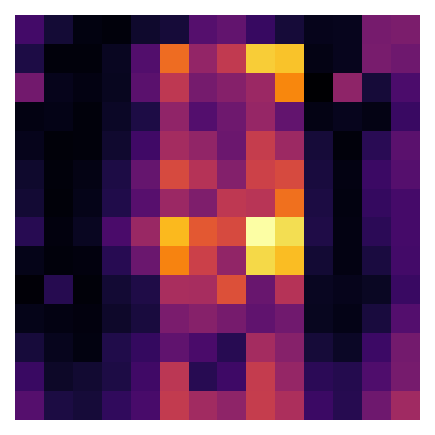} &
\includegraphics[width=0.13\linewidth]{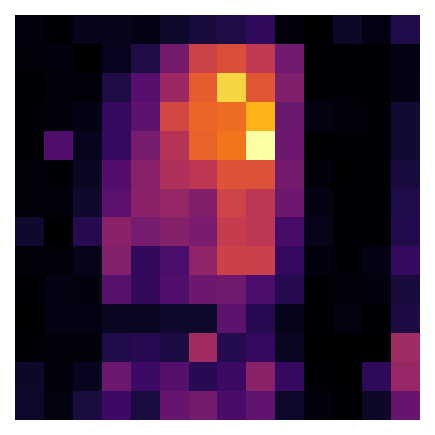} \\
[2pt]
\includegraphics[width=0.13\linewidth]{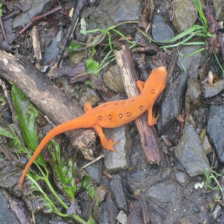} &
\includegraphics[width=0.13\linewidth]{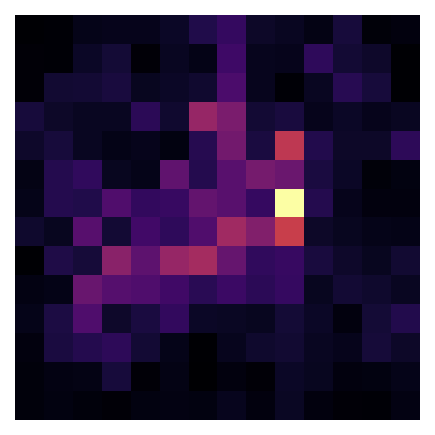} &
\includegraphics[width=0.13\linewidth]{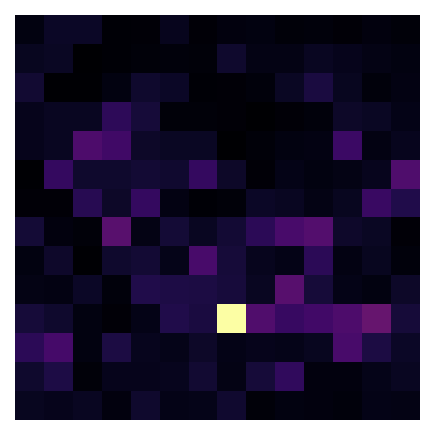} &
\includegraphics[width=0.13\linewidth]{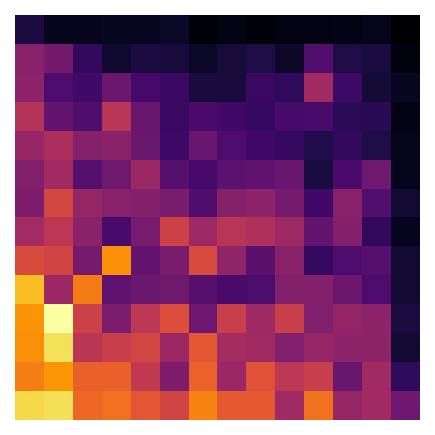} &
\includegraphics[width=0.13\linewidth]{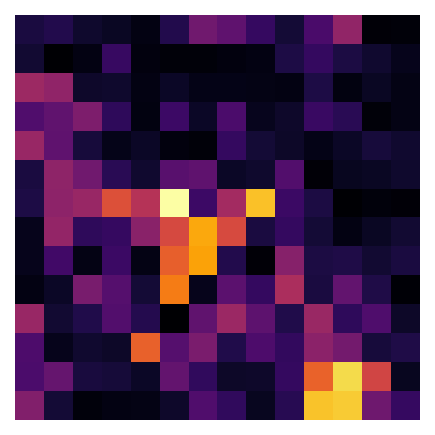} &
\includegraphics[width=0.13\linewidth]{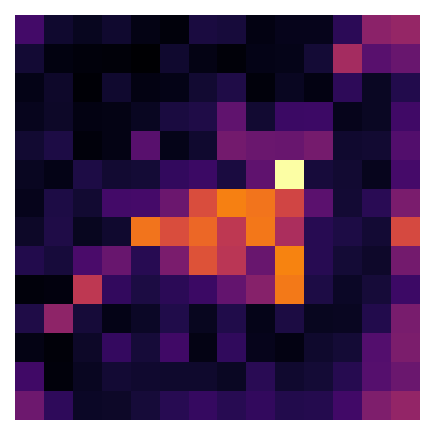} &
\includegraphics[width=0.13\linewidth]{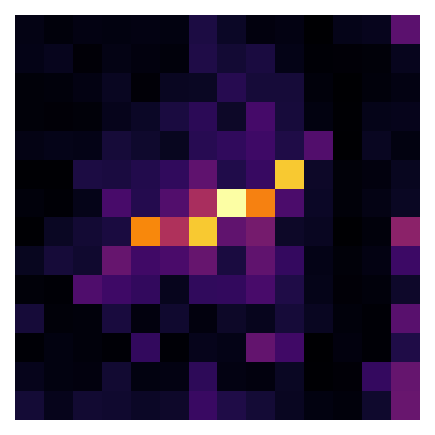} \\
[2pt]
\includegraphics[width=0.13\linewidth]{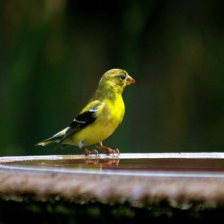} &
\includegraphics[width=0.13\linewidth]{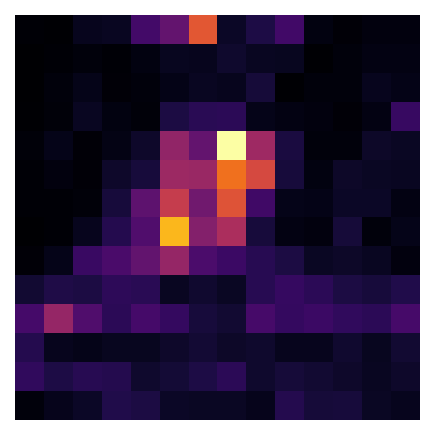} &
\includegraphics[width=0.13\linewidth]{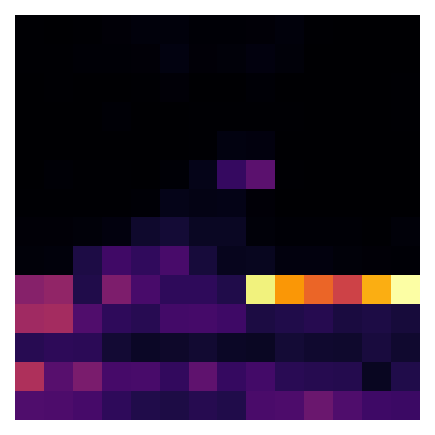} &
\includegraphics[width=0.13\linewidth]{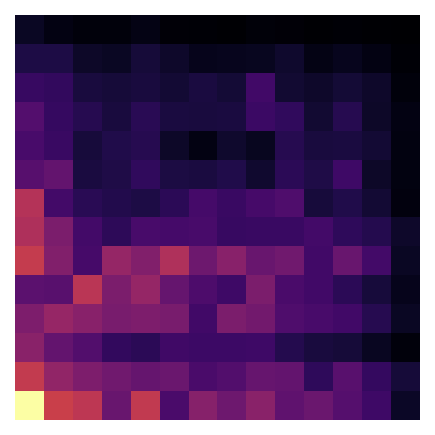} &
\includegraphics[width=0.13\linewidth]{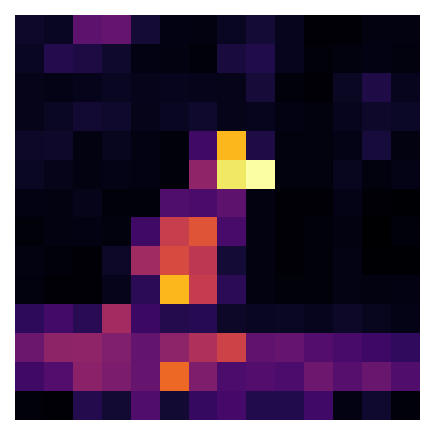} &
\includegraphics[width=0.13\linewidth]{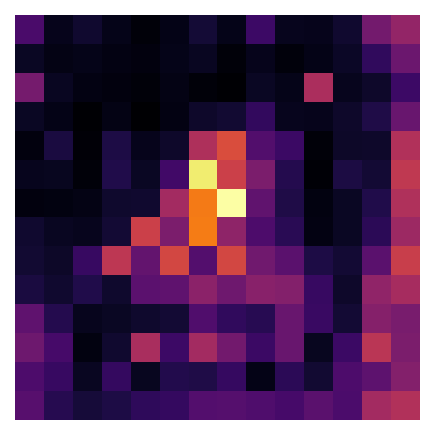} &
\includegraphics[width=0.13\linewidth]{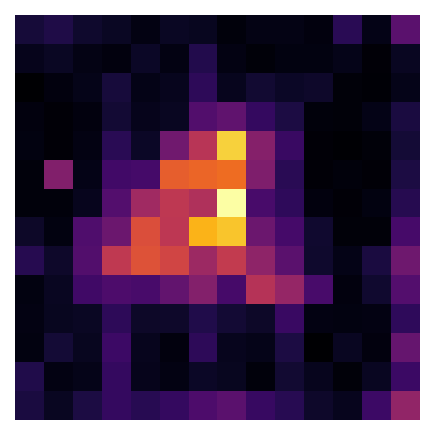} \\
[2pt]
\includegraphics[width=0.13\linewidth]{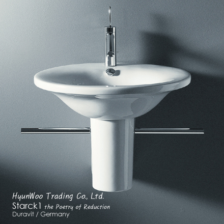} &
\includegraphics[width=0.13\linewidth]{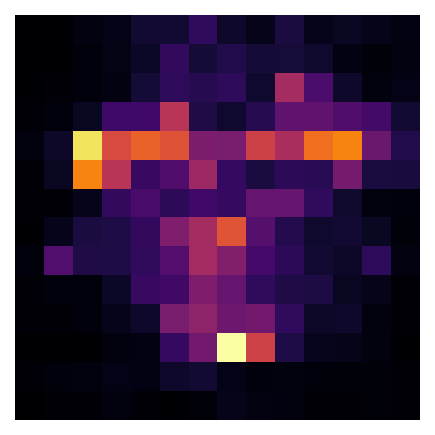} &
\includegraphics[width=0.13\linewidth]{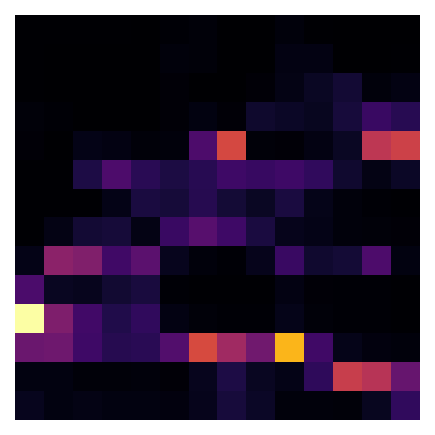} &
\includegraphics[width=0.13\linewidth]{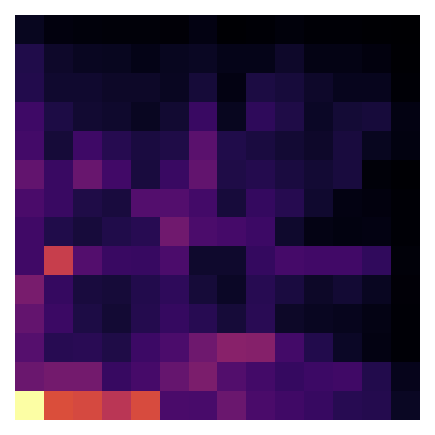} &
\includegraphics[width=0.13\linewidth]{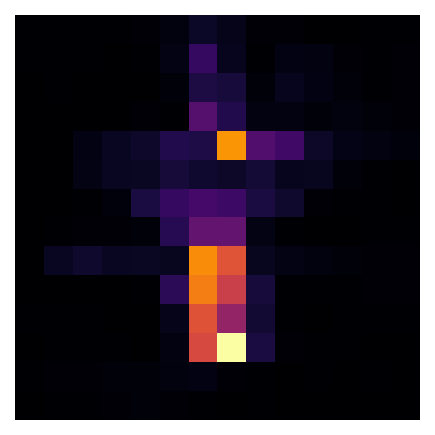} &
\includegraphics[width=0.13\linewidth]{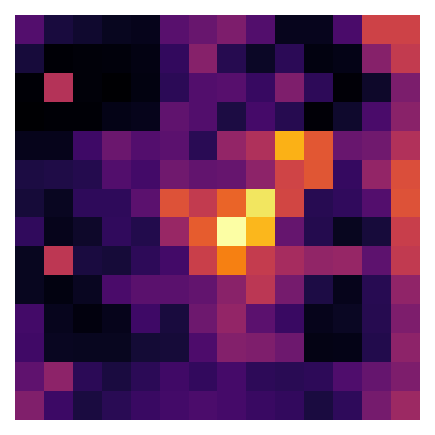} &
\includegraphics[width=0.13\linewidth]{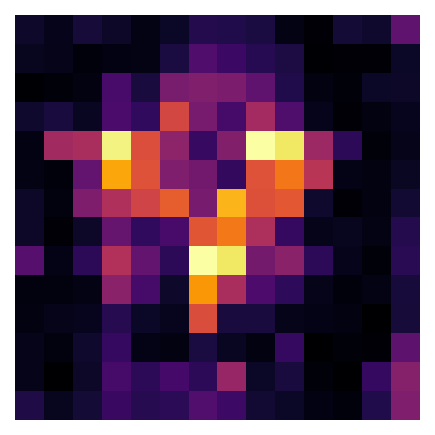} \\
\end{tabular}
\caption{{\bf Attention visualisations of encoder features (ViT-B/16).} Each row shows the original image followed by the self-attention map for DINO~\cite{caron2021dino} (400 epochs), iBOT~\cite{zhou2021ibot} (400 epochs), MAE~\cite{he2021mae} (1600 epochs), MSN~\cite{assran2022msn} (600 epochs), I-JEPA~\cite{assran2023ijepa} (600 epochs) and \model (600 epochs), all ViT-B/16 encoders pretrained on ImageNet-1K. All baseline visualisations are reproduced from publicly released checkpoints.}
\label{fig:att_vitb_appendix}
\end{figure*}

\end{document}